\documentclass[journal]{IEEEtran}
\ifCLASSINFOpdf
\else
\fi

\usepackage{amsmath}
\usepackage{amsthm}
\usepackage{algorithm}
\usepackage[noend]{algpseudocode}
\usepackage[psamsfonts]{amssymb}
\usepackage[pdftex]{graphicx}
\usepackage{tikz}
\usepackage{tabu}

\newtheorem{proposition}{Proposition}

\theoremstyle{definition}
\newtheorem{definition}{Definition}
\newtheorem{example}{Example}

\usetikzlibrary{shapes,arrows,automata}
\usetikzlibrary{shadows}
\usetikzlibrary{positioning}

\tikzstyle{plain} = [draw=none,fill=none]
\tikzstyle{decision} = [diamond, draw, top color=white, bottom color=blue!30, 
                            draw=blue!50!black!100, drop shadow,
    text width=1.5cm, text badly centered, inner sep=0pt]
\tikzstyle{block} = [rectangle, draw, top color=white, bottom color=blue!30, 
                            draw=blue!50!black!100, drop shadow,
    text width=2.5cm, text centered, rounded corners]
\tikzstyle{nodepolicy} = [circle, draw, top color=white, bottom color=blue!30, 
                            draw=blue!50!black!100, drop shadow,
    text width=.8cm, text centered]
\tikzstyle{inputpolicy} = [circle, draw, top color=white, bottom color=green!30, 
                            draw=green!50!black!100, drop shadow,
    text width=.8cm, text centered]
\tikzstyle{outputpolicy} = [circle,trapezium left angle=70,trapezium right angle=-70, draw, top color=white, bottom color=red!30, 
                            draw=red!50!black!100, drop shadow,
    text width=.8cm, text centered]
\tikzstyle{line} = [draw, -latex, ultra thick]
\tikzstyle{cloud} = [draw, ellipse,fill=red!20, node distance=3cm,
    minimum height=2em]

\newcommand{\mbs}[1]{\ensuremath{\boldsymbol{#1}}}

\newcommand{\ex}{\mathbb{E}}

\newcommand{\thetav}{\mbs{\theta}}

\newcommand{\kv}{\mathbf{k}}

\newcommand{\s}{\mathbf{s}}

\newcommand{\uv}{\mathbf{u}}
\newcommand{\x}{\mathbf{x}}
\newcommand{\y}{\mathbf{y}}

\newcommand{\I}{\mathbf{I}}
\newcommand{\K}{\mathbf{K}}

\newcommand{\X}{\mathbf{X}}

\newcommand{\at}{\mathbf{a}_t}

\newcommand{\atp}{\mathbf{a}_{t+1}}

\newcommand{\w}{\mathbf{w}}

\newcommand{\GP}{\mathcal{GP}}

\newcommand{\N}{\mathcal{N}}

\newcommand{\specialcell}[2][c]{%
  \begin{tabular}[#1]{@{}c@{}}#2\end{tabular}}

\begin{document}
%
\title{Funneled Bayesian Optimization for Design, Tuning and Control of Autonomous Systems}
%
%
%

\author{Ruben Martinez-Cantin
\thanks{R. Martinez-Cantin is at Centro Universitario de la Defensa, Zaragoza, Spain and SigOpt, Inc. email: rmcantin@unizar.es}}

%
%

\markboth{Journal of \LaTeX\ Class Files,~Vol.~14, No.~8, August~2015}%
{Shell \MakeLowercase{\textit{et al.}}: Bare Demo of IEEEtran.cls for IEEE Journals}
%



\maketitle

\begin{abstract}
In this paper, we tackle several problems that appear in robotics and autonomous systems: algorithm tuning, automatic control and intelligent design. All those problems share in common that they can be mapped to global optimization problems where evaluations are expensive. Bayesian optimization has become a fundamental global optimization algorithm in many problems where sample efficiency is of paramount importance. Bayesian optimization uses a probabilistic surrogate model to learn the response function and reduce the number of samples required. Gaussian processes have become a standard surrogate model for their flexibility to represent a distribution over functions. In a black-box settings, the common assumption is that the underlying function can be modeled with a stationary Gaussian process. In this paper, we present a novel kernel function specially designed for Bayesian optimization, that allows nonstationary behavior of the surrogate model in an adaptive local region. This kernel is able to reconstruct nonstationarity even with the irregular sampling distribution that arises from Bayesian optimization. Furthermore, tn our experiments, we found that this new kernel results in an improved local search (exploitation), without penalizing the global search (exploration) in many applications. We provide extensive results in well-known optimization benchmarks, machine learning hyperparameter tuning, reinforcement learning and control problems, and UAV wing optimization. The results show that the new method is able to outperform the state of the art in Bayesian optimization both in stationary and nonstationary problems. 
\end{abstract}


%
\IEEEpeerreviewmaketitle

\section{Introduction}

\IEEEPARstart{M}{any} problems in autonomous systems and robotics require to find the extremum of an unknown real valued function usign as few evaluations as possible. In many cases, those functions represent actual expensive processes like building a prototype, physical trials like learning a controller by experimentation, or time consuming computations and simulations like tuning some deep learning architecture. The optimization process must consider the actual budget and limitations of gathering new evaluations. Then, sample efficiency becomes the key element. Furthermore, those functions might be highly multimodal, requiring a global solution.

Bayesian optimization, also found in the literature with the names of Bayesian Sampling \cite{stuckman1992comparison}, Efficient Global Optimization (EGO) \cite{Jones:1998}, Sequential Kriging Optimization (SKO) \cite{Huang06}, Sequential Model-Based Optimization (SMBO) \cite{HutHooLey11-smac} or Bayesian guided pattern search \cite{taddy2009bayesian}, is a classic optimization method \cite{Kushner:1964,Mockus1989} which has become quite popular recently for being a sample efficient method of global optimization \cite{Jones:1998}. It has been applied with great success to autonomous algorithm tuning \cite{Mahendran:2012}, specially for machine learning applications \cite{Snoek2012,feurer2015efficient}, robot planning \cite{MartinezCantin09AR}, control \cite{Calandra2015a}, task optimization \cite{KroemerJRAS_66360}, reinforcement learning \cite{MartinezCantin07RSS,ActiveRewardLearning}, structural design \cite{forrester2006optimization}, sensor networks \cite{Srinivas10,Garnett2010}, simulation design \cite{Brochu:2010a}, circuit design \cite{taddy2009bayesian}, ecology \cite{wario2015automatic}, biochemistry \cite{czarnecki2015robust}, dynamical modeling of biological systems \cite{ulmasov2016bayesian}, etc. Recent works have found connections between Bayesian optimization and the way biological systems adapt and search in nature, such as human active search \cite{Borji2013} or animal adaptation to injuries \cite{Cully2015}.

Bayesian optimization uses a Bayesian surrogate model, that is, a distribution over target functions $P(f)$ to incorporate the information available during the optimization procedure, like any prior information and the information from each observation. This model can be updated recursively as outcomes are available from the evaluated trials $y_t = f(\x_t)$ 
\begin{equation}
  \label{eq:bayes-sec}
  P(f|\X_{1:t},\y_{1:t}) = \frac{P(\x_{t},y_{t}|f) P(f|\X_{1:t-1},\y_{1:t-1})}{P(\x_{t},y_{t})},
\end{equation}
$\forall \; t=2 \ldots T$ where $\X_{1:t}$ is a matrix with all the inputs $\mathbf{X}_{1:t}=[\x_1,\ldots,\x_t]$ and $\y_{1:t}$ is a vector with all the outcomes $\y_{1:t} = [y_1,\ldots,y_t]^T$. By using this method, the information is always updated. Therefore, the surrogate model provides a memory \cite{Moore:1996} that improves the sample efficiency of the method by considering the whole history of trials and evaluations during the decision procedure of where to sample.

Bayesian optimization computes the optimal decision/action $\uv$ of selecting the next trial $\uv = \x_{t+1}$ by maximizing (minimizing) a expected utility (loss):
\begin{equation}
\uv^{BO} = \arg \min_{\uv} \int \delta_t(f,\uv) \; dP(f|\X_{1:t},\y_{1:t})     
\label{eq:bayes-average}
\end{equation}
where $\delta_t(f,\uv)$ is the \emph{optimality criterion} or \emph{regret function} that drives the optimization towards the optimum $\x^*$. For example, we can use the \emph{optimality gap} $\delta_t(f,\uv) = f\left(\uv \right) - f(\x^*)$ to get the optimal outcome, the \emph{Euclidean distance error} $\delta_t(f,\uv) = \|\uv - \x^*\|_2$ to get the optimal input, or the \emph{relative entropy} $\delta_t(f,\uv) = H(\x^*|\X_{1:t}) - H(\x^*|\X_{1:t}, \uv)$ to maximize the information about the optimum.

In summary, Bayesian optimization is the combination of two main components: a surrogate model which captures all prior and observed information and a decision process which performs the optimal action, i.e.: where to sample next, based on the previous model. These two components also hide extra computational cost compared to other optimization methods. On one hand, we need to update the surrogate model continuously. On the other hand, we need to optimize the criterion function. However, this additional cost can be compensated by the reduced number of target function evaluations thanks to the sample efficiency of the method. Therefore, Bayesian optimization is specially suitable for expensive black-box functions, trial-and-error tuning and experimental design. The methodology behind Bayesian optimization also appears in other areas. In the way points are selected, the optimization problem can be considered an active learning problem on the estimation of the optimum \cite{HennigSchuler2012,deFreitas:2013}. Other authors have drawn connections between Bayesian optimization and some reinforcement learning setups such as multi-armed bandits \cite{Srinivas10}, partially observable Markov decision processes (POMDPs) \cite{ToussaintBSG} or active reinforcement learning \cite{MartinezCantin07RSS}. Surrogate models such as Gaussian processes are also used in other optimization methods, like evolutionary algorithms \cite{buche2005accelerating,cheng2015multiobjective,ong2003evolutionary}. Some genetic algorithms even use the decision (acquisition) functions used in Bayesian optimization as preselection criteria \cite{zhou2007combining}.

For Bayesian optimization, the quality of the surrogate model is of paramount importance as it also affects to the optimality of the decision process. Earliest versions of Bayesian optimization used Wiener processes \cite{Kushner:1964} or Wiener fields \cite{Mockus78} as surrogate models. Similar methods used radial basis functions \cite{gutmann2001radial} or branch and bound with polynomials \cite{sherali2003pseudo}. It was the seminal paper of Jones et al. \cite{Jones:1998} that introduced the use of Gaussian processes, also called Kriging, as a Bayesian surrogate function. Jones also wrote an excellent review on this kind of surrogate models \cite{Jones:2001}. Recently, other Bayesian models have become popular like Student's t processes \cite{AmarShah2014,MartinezCantin14jmlr}, treed Gaussian processes \cite{taddy2009bayesian,Assael2014}, random forests \cite{HutHooLey11-smac}, tree-structured Parzen estimators \cite{Bergstra2011} or deep neural networks \cite{snoek-etal-2015a}. In the case of discrete inputs, the Beta-Bernouilli bandit setting provides an equivalent framework \cite{russo2014learning}. 

However, the Gaussian process (GP) is still the most popular model due to its accuracy, robustness and flexibility, because Bayesian optimization is mainly used in black or grey-box scenarios. The range of applicability of a Gaussian process is defined by its kernel function, which sets the family of functions that is able to represent through the reproducing kernel Hilbert space (RKHS) \cite{Rasmussen:2006}. In fact, regret bounds for Bayesian optimization using Gaussian processes are always defined in terms of specific kernel functions \cite{deFreitas:2013,Srinivas10,Bull2011}. From a practical point of view, the standard procedure is to select a generic kernel function, such as the Gaussian (square exponential) or Mat{\'e}rn kernels, and estimate the kernel hyperparameters from data. One property of these kernels is that they are stationary. Although it might be a reasonable assumption in a black box setup, we show in Section \ref{sec:nst} that this reduces the efficiency of Bayesian optimization in most situations. It also limits the potential range of applications. On the other hand, nonstationay methods usually require extra knowledge of the function (e.g.: the global trend or the space partition). However, gathering this knowledge from data usually requires global sampling, which is contrary to the Bayesian optimization methodology.

The main contribution of the paper is a new set of adaptive kernels for Gaussian processes that are specifically designed to model functions from nonstationary processes but focused on the region near the optimum. Thus, the new model maintains the global exploration/local exploitation trade off. This idea results in an improved efficiency and applicability of any Bayesian optimization based on Gaussian processes. We call this new method \emph{Spartan Bayesian Optimization} (SBO). The algorithm has been extensively evaluated in many scenarios and applications. Besides some standard optimization benchmarks, the method has been evaluated in automatic algorithm tuning for machine learning applications, optimal policy learning in reinforcement learning scenarios and autonomous wing design of an airplane using realistic CFD simulations. In our results, we have found that SBO reaches its best performance in problems that are clearly nonstationary, where the local and global shape of the function are different. However, our tests have also shown that SBO can improve the results of Bayesian optimization in all those scenarios. From an optimization point of view, these new kernels result in an improved local search while maintaining global exploration capabilities, similar to other locally-biased global optimization algorithms \cite{gablonsky2001locally}. 

\section{Bayesian optimization with Gaussian processes}
\label{sec:bo}

We start describing the ingredients for a Bayesian optimization algorithm using Gaussian processes as surrogate model. Consider the problem of finding the minimum of an unknown real valued function $f:\mathbb{X} \rightarrow \mathbb{R}$, where $\mathbb{X}$ is a compact space, $\mathbb{X} \subset \mathbb{R}^d, d \geq 1$. In order to find the minimum, the algorithm has a maximum budget of $N$ evaluations of the target function $f$. The purpose of the algorithm is to select the best query points at each iteration such as the optimization gap or regret is minimum for the available budget. 


\subsection{Gaussian processes}
Given a dataset at step $t$ of points $\X=\X_{1:t}$ and its respective outcomes $\y=\y_{1:t}$, the prediction of the Gaussian process at a new query point $\x_q$, with a kernel or covariance function $k_{\thetav}(\cdot,\cdot)$ with hyperparameters $\thetav$ is a normal distribution such as $y_q \sim \N(\mu,\sigma^2|\x_q)$ where:
\begin{equation}
  \begin{split}
\label{eq:predgp}
\mu(\x_q) &= \kv(\x_q,\X) \K^{-1} \y  \\
\sigma^2(\x_q) &= k(\x_q,\x_q) - \kv(\x_q,\X) \K^{-1} \kv(\X,\x_q)    
  \end{split}
\end{equation}
being $\kv(\x_q,\X)$ the corresponding cross-correlation vector of the query point $\x_q$ with respect to the dataset $\X$
\begin{equation}
  \kv(\x_q,\X) = \left[  k(\x_q,\x_1), \ldots,  k(\x_q,\x_n) \right]^T
  \label{eq:2}
\end{equation}
and $\K = \K(\X,\X)$ is the Gram or kernel matrix:
\begin{equation}
\K  = \left(
\begin{array}{ccc}
  k(\x_1,\x_1)& \ldots & k(\x_1,\x_n)\\
  \vdots& \ddots & \vdots\\
  k(\x_n,\x_1)& \ldots & k(\x_n,\x_n)
\end{array}
\right) + \sigma^2_n\I
\label{eq:km}
\end{equation}
where $\sigma^2_n$ is a noise or nugget term to represent stochastic functions \cite{Huang06} or surrogate missmodeling \cite{Gramacy2012}.

\subsection{Acquisition function}
The regret functions that we introduced to select the next point at each iteration with equation (\ref{eq:bayes-average}) assume knowledge of the optimum $\x^*$. Thus, they cannot be used in practice. Instead, the Bayesian optimization literature uses \emph{acquisition functions}, like the \emph{expected improvement} (EI) criterion \cite{Mockus1989} as a proxy of the optimality gap criterion. EI is defined as the expectation of the improvement function $I(\x) = \max(0,\rho - f(\x))$, where $\rho$ is the incumbent of the optimal. In many applications $\rho$ is considered to be the \emph{best outcome} until the current iteration $\rho = y_{best}$. Other incumbent values could be considered, specially in the presence of noise, like the maximum predicted value. Taking the expectation over the predictive distribution, we can compute the expected improvement as:
\begin{equation}
  \label{eq:eigen}
  EI(\x) = \ex_{p(y|\x,\thetav)} \left[I(\x)\right] = \left(\rho - \mu\right) \Phi(z) + \sigma \phi(z)
\end{equation}
where $\phi$ and $\Phi$ are the corresponding Gaussian probability density function (PDF) and cumulative density function (CDF), being $z = (\rho - \mu)/\sigma$. In this case, $(\mu,\sigma^2)$ are the prediction parameters computed with Equation (\ref{eq:predgp}). At each iteration $n$, we select the next query at the point that maximizes the corresponding expected improvement:
\begin{equation}
\x_n = \arg \max_{\x} \; EI(\x)
\label{eq:ei}  
\end{equation}

\subsection{Kernel hyperparameters}
The advantage of Gaussian process is that the posterior can be computed in closed form due to the linear-Gaussian properties of all the components involved. This is true for known kernel hyperparameters. Uncertainty in the hyperparameters requires a nonlinear transformation which makes the computation of the predictive distribution or the acquisition function intractable. The most common approach is to use a point estimate of the kernel hyperparameters based on the maximum of the marginal likelihood $p(\y | \X, \thetav)  = \N(0, \K)$ \cite{Rasmussen:2006}. Sometimes a prior on the hyperparameters $p(\thetav)$ is defined, resulting in a maximum a posteriori point estimate.

Instead, we propose a fully-Bayesian treatment, where we compute the integrated predictive distribution and the integrated acquisition function:
\begin{equation}
  \label{eq:intpost}
  \begin{split}
    \widehat{y_q} &= \int \N(\mu, \sigma | \x_q) p(\thetav | \y, \X) d\thetav\\
    \widehat{EI}(\x) &= \int EI(\x) p(\thetav | \y, \X) d\thetav
  \end{split}
\end{equation}
with respect to the posterior distribution on the hyperparameters $p(\thetav | \y, \X) \propto p(\y | \X, \thetav) p(\thetav)$. Those integrals are directly intractable, thus, we use Markov chain Monte Carlo (MCMC) to generate a set of samples $\mathbf{\Theta} = \{\thetav_i\}_{i=1}^N$ with $\thetav_i \sim p(\thetav | \y, \X)$. We use the slice sampling algorithm which has already been used in Bayesian optimization \cite{Snoek2012}, although bootstraping could equally be used \cite{kleijnen2012expected}. Compared to point estimates of $\thetav$ \cite{Jones:1998}, MCMC has a higher computational cost, but MCMC has shown to be more robust \cite{Snoek2012,MartinezCantin14jmlr} in Bayesian optimization. Note that, because we use a sampling distribution of $\thetav$ the predictive distribution at any point $\x$ is a sum of Gaussians, that is:
\begin{equation}
  \label{eq:mcmcpost}
  \begin{split}
    \widehat{y_q} &= \sum_{i=1}^N \N(\mu_i, \sigma_i | \x_q)\\
    \widehat{EI}(\x) &= \sum_{i=1}^N \left(\rho - \mu_i\right) \Phi(z_i) + \sigma_i \phi(z_i)
  \end{split}
\end{equation}
where each $(\mu_i, \sigma_i, z_i)$ is computed using the $i$-th sample of the kernel hyperparameters $\thetav_i$ using Equation \eqref{eq:predgp}.

\subsection{Initialization}
Finally, in order to avoid bias and guarantee global optimality, we rely on an initial design of $p$ points based on \emph{Latin Hypercube Sampling} (LHS) following the recommendation in the literature \cite{Jones:1998,Bull2011,kandasamy2015high}. Algorithm \ref{al:bo} summarizes the basic steps in Bayesian optimization.

\begin{algorithm}
\renewcommand{\algorithmicrequire}{\textbf{Input:}}
\caption{Bayesian optimization (BO)}\label{al:bo}
\begin{algorithmic}[1]
\Require Total budget $T$, initialization budget $p$.
\State $\X \gets \x_{1:p} \;\;\;  \y \gets y_{1:p}$ \Comment{Initial design with LHS}
\For{$t = p \ldots T$}                \Comment{Available budget of $N$ queries}
   \State $\mathbf{\Theta} \gets \texttt{SampleHyperparameters}(\X,\y)$  
   \State $\x_t = \arg \max_{\x} \; \widehat{EI}(\x | \X,\y,\mathbf{\Theta})$ from Eq. \eqref{eq:mcmcpost}
   \State $y_t \gets f(\x_t) \qquad \X \gets\texttt{add}(\x_{t}) \qquad  \y \gets \texttt{add}(y_{t})$
\EndFor
\end{algorithmic}
\end{algorithm}


\section{Nonstationary Gaussian processes}
\label{sec:nst}

Many applications of Gaussian process regression, including Bayesian optimization, are based on the assumption that the process is stationary. This is a reasonable assumption for black-box optimization as it does not assume any extra information on the evolution of the function in the space.

\begin{definition}
Let $f(\x) \sim \GP(0, k(\x,\x'))$ be a zero mean Gaussian process. We say that $f(\x)$ is a \emph{stationary} process if the kernel function  $k(\x,\x')$ is stationary, that is, it can be written as $k(\x,\x') = k(\mathbf{\tau})$ where $\mathbf{\tau} = \x-\x'$.
\end{definition}
This is equivalent to say that the process is invariant to translation, that is, the statistical properties of the set of points $\{\x_1, \ldots, \x_n\}$ are the same as for the points $\{\x_1 + h, \ldots, \x_n + h\}$ for any real vector $h$. In practice, a process is only used for limited distances. For example, in our definition of Bayesian optimization, we already limited our search for the minimum to a compact space $\mathbb{X}$. Thus, if we define $b$ as the diameter or the longest distance enclosed in $\mathbb{X}$, then, for practical reasons, the process is stationary if it is invariant for $|h| \leq b$. In other circumstances, we might find that our process is translation invariant in a smaller region $\mathbb{X}_c \subset \mathbb{X}$, with diameter $b_c$. In this case, we say that the process is \emph{locally stationary} for $|h| \leq b_c$. In the geostatistics community, this is also called \emph{quasi-stationarity} \cite{Journel78}. Note that a locally stationary process is also stationary in any smaller subset $\mathbb{X}_s \subseteq \mathbb{X}_c$. Intuitively, even for nonstationary process, the smaller the region defined by distance $b_c$, the more homogeneous would be process relatively to larger regions, being closer to locally stationary \cite{HAAS19901759}. Finally, any locally stationary process is \emph{nonstationary} for any set $\mathbb{Y} \subseteq \mathbb{X}$ which contains elements outside the locally stationary region $\exists \; \x_f \in \mathbb{Y}, \x_f \notin \mathbb{X}_c$.

Bayesian optimization is a locally homogeneous process. For most acquisition functions, it has a dual behavior of global exploration and local exploitation. Typically, both sampling and uncertainty estimation end up being distributed unevenly, with many samples and small uncertainty near the local optima and sparse samples and large uncertainty everywhere else. Furthermore, many direct applications of Bayesian optimization are inherently nonstationary. Take for example the reinforcement learning problems of Section \ref{sec:reinf-learn}:
\begin{example}
Let us consider a biped robot (agent) trying to find the walking pattern (policy) that maximizes the walking speed (reward). In this setup, there are some policies that reach undesirable states or result in a failure condition, like the robot falling or losing the upright posture. Then, the system returns a null reward or arbitrary penalty. In cases where finding a stable policy is difficult, the reward function may end up being almost flat, except for a small region of successful policies where the reward is actually informative in order to maximize the speed.
\end{example}
The reward function is directly nonstationary for this dual behaviour between failure and success. Modeling this kind of functions with Gaussian processes require kernels with different length scales for the flat/non-flat regions or specially designed kernels to capture that behavior. 

There has been several attempts to model nonstationary functions with Gaussian processes. For example, the use of specific nonstationary kernels \cite{Rasmussen:2006}, space partitioning \cite{krause07nonmyopic}, Bayesian treed GP models \cite{gramacy2005bayesian} or projecting (warping) the input space to a stationary latent space \cite{sampson1992nonparametric}. The idea of treed GPs was used in Bayesian optimization combined with an auxiliary local optimizer \cite{taddy2009bayesian}. A version of the warping idea was applied to Bayesian optimization \cite{snoek-etal-2014a}. Later, Assael et al. \cite{Assael2014} built a treed GPs where the warping model was used in the leaves. These methods try to model nonstationarity in a global way. For Bayesian optimization the best fit for a global model might end up being inaccurate near the minimum, where we require more accuracy.

Before explaining our approach to nonstationarity, we are going to review the most common stationary kernels and the effect of the hyperparameters on the optimization process. More details can be found in Appendix~\ref{ax:length}.

\subsection{Stationary kernels}
Most popular stationary kernels include the squared exponential (SE) kernel and the Mat{\'e}rn kernel family:
\begin{align}
  k_{SE}(\x,\x') &= \exp\left(-\frac{1}{2} r^2\right) \label{eq:se}\\
  k_{M,\nu}(\x,\x') &= \frac{2^{1-\nu}}{\Gamma(\nu)}\left(\sqrt{2\nu r}\right)^\nu K_\nu\left(\sqrt{2\nu r}\right)
\end{align}
where $r^2= (\x-\x')^T \Lambda (\x-\x')$ with some positive semidefinite $\Lambda$. More frequently, $\Lambda = diag(\thetav_l^{-1})$ where $\thetav_l$ represents the length-scale hyperparameters that capture the smoothness or variability of the function. If $\thetav_l$ is a scalar, the function is isotropic. If $\thetav_l$ is a vector we can estimate a different smoothness per dimension, which also represents the relevance of the data in that dimension. Thus, the later case is called automatic relevance determination (ARD) \cite{Rasmussen:2006}.

In the Mat{\'e}rn kernel, $K_\nu$ is a modified Bessel function \cite{Rasmussen:2006}. The Mat{\'e}rn kernel is usually computed for values of $\nu$ that are half-integers $\nu=p+1/2$ where $p$ is a non-negative integer, because the function becomes simpler. For example:
 \begin{align}  
  k_{M,\nu=1/2}(\x,\x') &= \exp\left(-r\right)\\
  k_{M,\nu=3/2}(\x,\x') &= \exp\left(-\sqrt{3} r\right) \left(1+\sqrt{3} r\right)\\
  k_{M,\nu=5/2}(\x,\x') &= \exp\left(-\sqrt{5} r\right) \left(1+\sqrt{5} r + \frac{5}{3} r^2\right)   \label{eq:matern}
\end{align}
The value of $\nu$ is related to the smoothness of the functions because it represents the $\eta$th-differentiability of the underlying process. That is, the process $g(x)$ defined by the Mat{\'e}rn kernel $k(\cdot,\cdot)$ is $\eta$-times mean square differentiable if and only if $\nu > \eta$ \cite{Rasmussen:2006}. Furthermore, for $\nu\rightarrow \infty$, we obtain the SE kernel from equation \eqref{eq:se}. In Bayesian optimization, the Mat{\'e}rn kernel with $\nu = 5/2$ is frequently used for its performance in many benchmarks, because it can properly represents smooth and irregular functions without imposing excessive restrictiveness like the infinite differentiability of the SE kernel.

\subsubsection{The effect of length-scales}
\label{sec:length-scales}
For optimization, the analysis of the kernel length-scale plays an important role for optimization. Thus, \emph{having multiple length-scales in a single process opens new possibilities besides modeling nonstationarity}. Small values of $\theta_l$ will be more suitable to capture signals with high frequency components; while large values of $\theta_l$ result in a model for low frequency signals or flat functions. Therefore, for the same distance between points, a kernel with smaller length-scale will result in higher predictive variance, therefore the exploration will be more aggressive. This idea was previously explored in Wang et al. \cite{ZiyuWang2013} by forcing smaller scale parameters to improve the exploration. More formally, in order to achieve no-regret convergence to the minimum, the target function must be an element of the reproducing kernel Hilbert space (RKHS) characterized by the kernel $k(\cdot,\cdot)$ \cite{Bull2011,Srinivas10}. For a set of kernels like the SE or Mat{\'e}rn, it can be shown that:
\begin{proposition}
Given two kernels $k_l$ and $k_s$ with large and small length scale hyperparameters respectively, any function $f$ in the RKHS characterized by a kernel $k_l$ is also an element of the RKHS characterized by $k_s$ \cite{ZiyuWang2013}.  
\end{proposition}
Thus, using $k_s$ instead of $k_l$ is safer in terms of guaranteeing no-regret convergence. However, if the small kernel is used everywhere, it might result in unnecessary sampling of smooth areas. Instead, we should combine both behaviors. The idea of combining kernels with different length-scales was previously hinted as a potential advantage for Bayesian optimization \cite{HennigSchuler2012}, but this is the first work to exploit its advantages in practice.

\subsection{Nonstationary combined kernels}
Our approach to nonstationarity is based on previous works where the input space is partitioned in different regions \cite{krause07nonmyopic,Nott2002}. The resulting GP is the linear combination of local GPs: $\xi(\x) = \sum_j \lambda_j (\x) \xi_j(\x)$. This approach was inspired by the concept of \emph{quasi-stationarity} from geostatistics \cite{Journel78,HAAS19901759}. This is equivalent to having a single GP with a combined kernel function of the form: $k(\x,\x'| \thetav) = \sum_{j} \lambda_j (\x) \lambda_j (\x') k_j(\x,\x'|\thetav_j)$. The final kernel is a valid kernel function as the sum and vertical scaling of kernels are also valid kernel functions \cite{Rasmussen:2006}. Each local kernel $k_j$ has its own specific hyperparameters $\thetav_j$ and statistical properties associated with the region, making the final kernel nonstationary even if the local kernels are stationary. A related approach of additive GPs was used by Kandasamy et al. to optimize high dimensional functions under the assumption that the actual function is a combination of lower dimensional functions \cite {kandasamy2015high}


\section{Spartan Bayesian Optimization}
\begin{figure}
  \centering
  \includegraphics[width=0.70\linewidth]{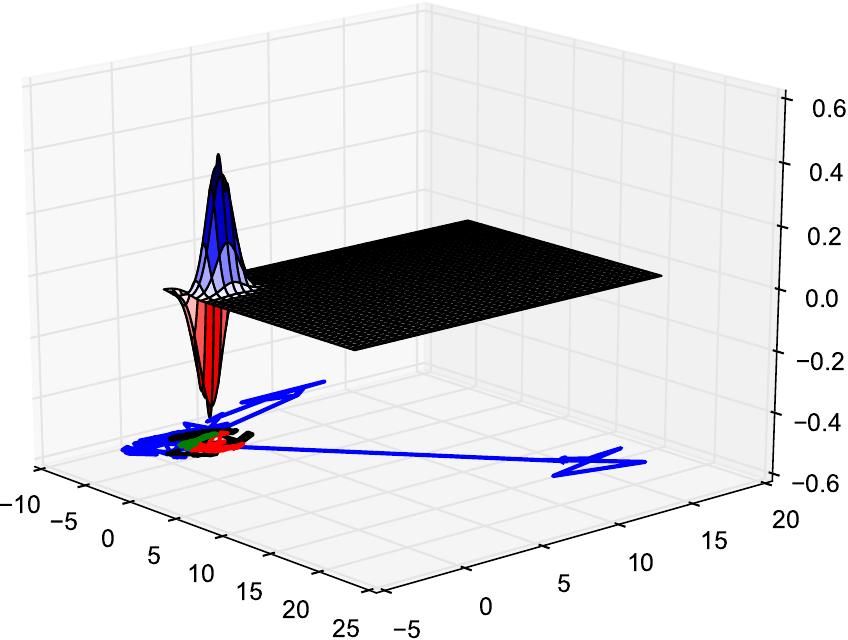}
  \caption{Left: Exponential 2D function from Gramacy \cite{gramacy2005bayesian}. The path below the surface represents the location of the local kernel as being sampled by MCMC. Clearly, it converges to the nonstationary section of the function. For visualization, the path is broken in colors by the order in the path (blue $\rightarrow$ black $\rightarrow$ green $\rightarrow$ red).}
  \label{fig:exp2dres}
\end{figure}

We present our new method, SBO, which combines a set of moving local and global kernels. This allows different control of the kernel length-scales in a local and global manner. Therefore, it is intrinsically  able to deal with nonstationarity, but it also allows different local and global behaviors for optimization as discussed in Section \ref{sec:length-scales}. For example, it can be used to improve local exploration towards the optimum without increasing global exploration. The intuition behind SBO is the same of many acquisition functions in Bayesian optimization: \emph{the aim of the surrogate model is not to approximate the target function precisely in every point, but to provide information about the location of the minimum}. Many optimization problems are difficult due to the fact that the region near the minimum is different from the rest of the function, e.g.: it has higher variability than the rest of the space, like the function in Fig. \ref{fig:exp2dres}. Also, it allows a better estimate of the hyperparameters of the local kernels without resulting in overconfidence of the hyperparameters globally. In those cases, SBO greatly improves the performance of the state of the art in Bayesian optimization.

\begin{figure}
  \centering
  \includegraphics[width=0.7\linewidth]{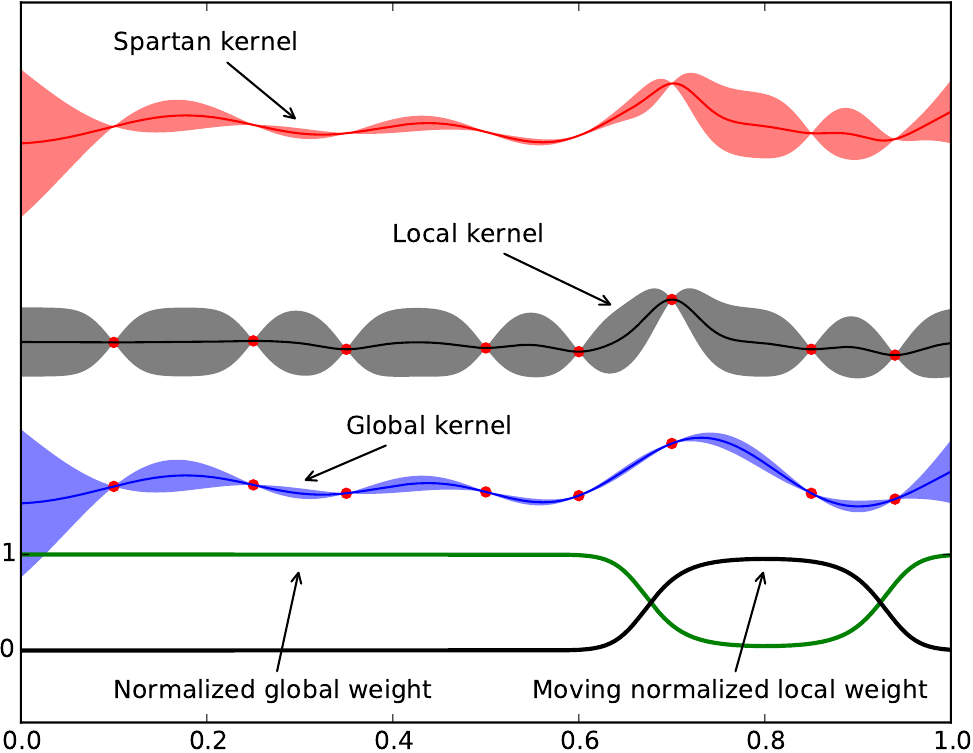}
  \caption{Representation of the Spartan kernel in SBO. In this case, the kernel is just a combination of a single local and global kernels. Typically, the local and global kernels have a small and large length-scale respectively. The influence of each kernel is represented by the normalized weight at the bottom of the plot. Note how the kernel with small length-scale produce larger uncertainties which is an advantage for fast exploitation, but it can perform poorly for global exploration as it tends to sample equally everywhere. On the other hand, the kernel with large length-scale provides a better global estimate, which improves focalized global exploration.}
  \label{fig:domains}
\end{figure}

We propose the combination of a kernel with global influence with several kernels with moving local influence. The influence is determined by a weighting function (see Figure \ref{fig:domains}). The influence of the local kernels is centered in a single point with multiple diameters, creating a funnel structure. We have called this kernel, the Spartan kernel:
\begin{equation}
  \label{eq:spartan}
  \begin{split}
k_{S}(\x, \x'|& \thetav_{S}) = \lambda_{g}(\x) \lambda_{g}(\x') k_{g}(\x,\x'| \thetav_g) \\ &+ \sum_{l=1}^M \lambda_l(\x| \thetav_{p}) \lambda_l(\x'| \thetav_{p}) k_l(\x,\x'| \thetav_l)
  \end{split}
\end{equation}
where the weighting function for the local kernel $\lambda_{l}(\x| \thetav_{p})$ includes the parameters to move the center of the local kernels along the input space. In order to achieve smooth interpolation between regions, each region have an associated weighting function $\omega_j(\x)$, having the maximum in the corresponding region $j$ and decreasing its value with distance to region $j$ \cite{Nott2002}. Then, we can set $\lambda_j(\x) = \sqrt{\omega_j(\x)/\sum_p \omega_p(\x)}$. The unnormalized weights $\omega$ are defined as:
\begin{equation}
  \label{eq:weights}
  \begin{aligned}
    \omega_{g} &= \N\left(\psi, \I \sigma^2_{g} \right) \\
    \omega_{l} &=  \N\left(\thetav_{p}, \I \sigma^2_{l} \right) \qquad \forall \; l = 1\ldots M
  \end{aligned}
\end{equation}
where $\psi$ and $\thetav_{p}$ can be seen as the center of the influence region of each kernel while $\sigma_{g}$ and $\sigma_{l}$ are related to the diameter of the area of influence. Note that all the local kernels share the same position (mean value) but different size (variance), generating a funnel-like structure. The Spartan kernel with a single local kernel can be seen in Fig. \ref{fig:domains}.

\paragraph{Global kernel parameters}
Unless we have prior knowledge of the function, the parameters of the global kernel are mostly irrelevant. In most applications, we can use a uniform distribution, which can be easily approximated by a large $\sigma^2_{g}$. For example, assuming a normalized input space $\mathcal{X} = [0,1]^d$, we can set $\psi = [0.5]^d$ and $\sigma^2_{g} = 10$.

\paragraph{Local kernel parameters}
For the local kernels, we estimate the center of the funnel structure $\thetav_p$ based on the data gathered. We propose to consider $\thetav_p$ as part of the hyperparameters for the Spartan kernel, which also includes the parameters of the local and global kernels, that is,
\begin{eqnarray}
  \label{eq:spakernel}
\thetav_{S}=[\thetav_{g}, \thetav_{l_1},\ldots,\thetav_{l_M}, \thetav_{p}]
\end{eqnarray}
The area of influence of each local kernel could also be adapted including the terms $\{\sigma^2_{l}\}_{l = 1}^M$ as part of the kernel hyperparameters $\thetav_{S}$. In practice, we found that the algorithm was learning the trivial cases of very small values of $\sigma^2_{l}$ in many experiments. As discussed in Section \ref{sec:nst}, for a small enough region, the behavior is stationary. However, a very small region ends up with not enough data points to properly learn the length-scale hyperparameters. Haas \cite{HAAS19901759} used a automatic sizing strategy by combining multiple heuristic and a minimum number of samples inside. Because we use a funnel structure, we found simpler to fix the value of $\sigma^2_{l}$ at different sizes. As can be seen in Section \ref{sec:results}, this simple heuristic provide excellent results. Alternatively, we can define $\sigma^2_{l}$ in terms of a fixed numbers of samples inside. This method has the advantage that, while doing exploitation, as the number of local samples increases, the funnel gets narrower, allowing better local refinement. 

As commented in Section \ref{sec:bo}, when new data is available, all the parameters are updated using MCMC. Therefore, the position of the local kernel $\thetav_p$ is moved each iteration to represent the posterior. Due to the sampling behavior in Bayesian optimization, we found that it intrinsically moves more likely towards the more densely sampled areas in many problems, which corresponds to the location of the function minima. Furthermore, as we have $N$ MCMC samples, there are $N$ different positions for the funnel of local kernels. On the other hand, for ARD local and global kernels with one length-scale per dimension $d$, the size of $\thetav_S$ is $(M+2)d$. Thus, the cost of the MCMC is $\mathcal{O}(N (M+2) d)$. Nevertheless, this cost is still cheaper than alternative methods (see Section \ref{sec:time}).

Spartan Bayesian Optimization is simple to implement from standard Bayesian optimization: it only requires to modify the kernel function and hyperparameter estimation. However, as we will see in Section \ref{sec:results}, the results show a large gain in terms of convergence and sample efficiency for many problems. 

It is important to note that, although we have implemented SBO relying on Gaussian processes and \emph{expected improvement}, the Spartan kernel also works with other popular kernel-based models such as Student-t processes \cite{Williams_Santner_Notz_2000,O'Hagan1992}, treed Gaussian processes \cite{gramacy2005bayesian} and other criteria such as \emph{upper confidence bound} \cite{Srinivas10}, \emph{relative entropy} \cite{NIPS2014_5324,HennigSchuler2012}, etc. A similar approach of using a moving local kernel was used in geostatics previously for terrain modeling \cite{HAAS19901759}. In that case, the local kernel was always centered in the query point, creating a sliding window approach to prediction. Therefore, even though the model was nonstationary, there was no different behaviour in different regions of the space.

\section{Evaluation and results}
\label{sec:results}

We have selected a variety of benchmarks from different applications to test our method. The purpose is twofold. First, we highlight the potential applicability of Bayesian optimization in many ways in autonomous systems, from design, control, software tuning, etc. Second, we show that in all those setups, our method outperforms the state of the art in Bayesian optimization.
\begin{enumerate}
\item We have taken several well-known functions for testing global optimization algorithms. The set of functions includes both stationary and non-stationary problems.
\item For the algorithm tuning and perception problems, we have selected a set of machine learning tuning benchmarks. They include large image classification and clustering problems.
\item For the control/reinforcement learning problems, we have selected some classic problems and a highly complex benchmark to control a hovering aerobatic helicopter.
\item Finally, we address the automatic design and embodiment problem with the optimal design of a wing profile for a UAV. This is a highly complex scenario, due to the chaotic nature of fluid dynamics. Thus, this problem is ubiquitous in global optimization and evolutionary computation literature.
\end{enumerate}

\subsection{Implementation details}
For evaluation purposes and to highlight the robustness of SBO, we have simplified the funnel structure to a single local and global kernel as in Fig. \ref{fig:domains}. We also took the simpler approach to fix the variance of $\omega_l$. We found that a single value of $\sigma^2_{l} = 0.05$ was robust enough in all the experiments once the input space was normalized to the unit hypercube.

For the experiments reported here we used a Gaussian process with unit mean function like in \cite{Jones:1998}. Although SBO allows for any combination of local and global kernels, for the purpose of evaluation, we used the Mat{\'e}rn kernel from equation (\ref{eq:matern}) with automatic relevance determination for both --local and global-- kernels. Furthermore, the kernel hyperparameters were initialized with the same prior for the both kernels. Therefore, we let the data determine which kernel has smaller length-scale. We found that the typical result is the behavior from Fig. \ref{fig:domains}. However, in some problems, the method may learn a model where the local kernel has a larger length-scale (i.e.: smoother and smaller variance) than the global kernel, which may also improve the convergence in plateau-like functions. Besides, if the target function is stationary, the system might end up learning a similar length-scale for both kernels, thus being equivalent to a single kernel. We can say that standard BO is a special case of SBO where the local and global kernels are the same. 

Given that for a single Mat{\'e}rn kernel with ARD, the number of kernel hyperparameters is the dimensionality of the problem, $d$, the number of hyperparameters for the Spartan kernel in this setup is $3d$. As we will see in the experiments, this is the only drawback of SBO compared to standard BO, as it requires running MCMC in a larger dimensional space, which results in higher computational cost. However, because SBO is more efficient, the extra computational cost can be easily compensated by a reduced number of samples. For the MCMC part, we used the slice sampling algorithm with a small number of samples (10), while trying to decorrelate every resample with larger burn-in periods (100 samples) as in Snoek et al. \cite{Snoek2012}.


We implemented Spartan Bayesian Optimization (SBO) using the BayesOpt library \cite{MartinezCantin14jmlr} as codebase. For comparison, we also implemented the input warping from Snoek et al. \cite{snoek-etal-2014a}, which we called WARP. To our knowledge, this is the only Bayesian optimization algorithm that has dealt with nonstationarity using Gaussian processes in a fully correlated way. However, as presented in Section \ref{sec:time}, WARP is much more expensive than SBO in terms of computational cost. For the warping function we used the cumulative density function of a Beta distribution or Beta CDF. The $(\alpha,\beta)$ parameters of the Beta CDF were also computed using MCMC.

We also did some preliminary tests with random forests (RF) \cite{HutHooLey11-smac} and treed GPs \cite{taddy2009bayesian,Assael2014}, which for clarity we have not included in the plots. For RF, we verified previous results in the literature which show their poor performance for continuous variables compared to GPs \cite{MartinezCantin14jmlr,Eggensperger2015}. For treed GPs, we found that they required a larger nugget or noise term $\sigma^2_n$ to avoid numerical instability. This is primarily due to the reduced number of samples per leaf and reduced global correlation. Increasing the noise term reduced considerably the accuracy of the results. Furthermore, single nonstationary GPs like the input warping method or our Spartan kernel can be used as tree leaves \cite{Assael2014}. Thus, our method can be used as an enhancement of treed GPs.

\begin{figure*}
  \centering
  \includegraphics[width=0.32\linewidth]{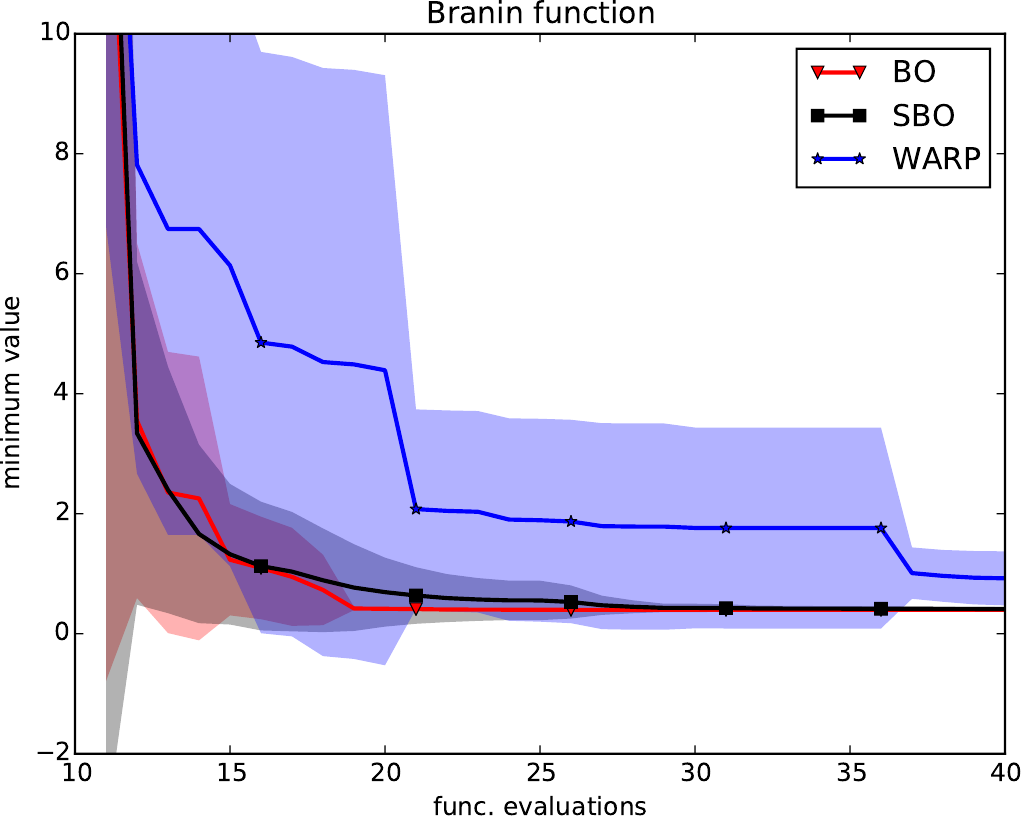}
  \includegraphics[width=0.32\linewidth]{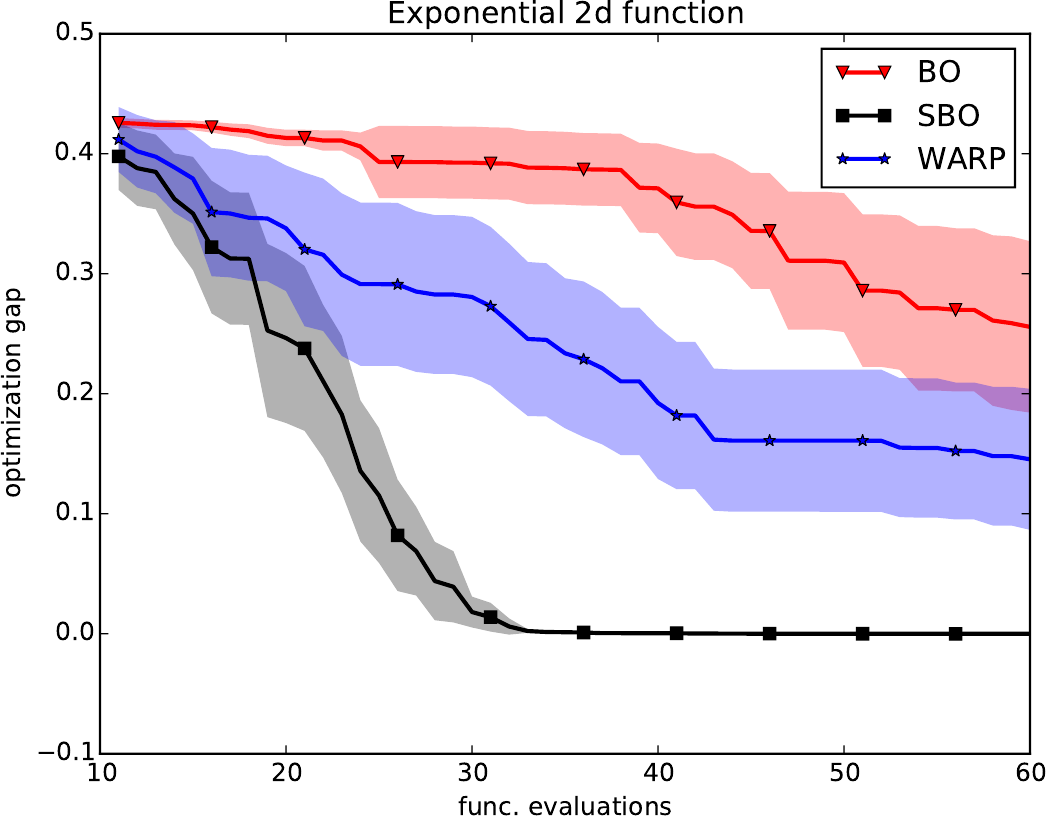}
  \includegraphics[width=0.32\linewidth]{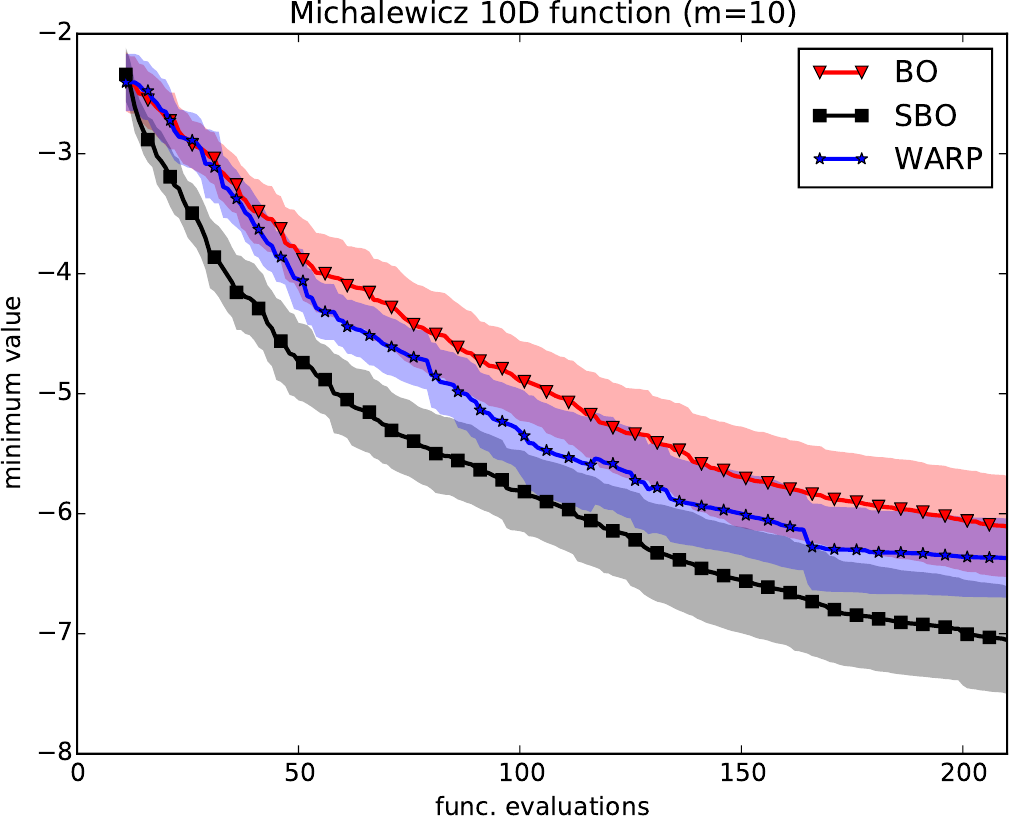}
  \caption{Optimization benchmarks. From left to right: Results on the Branin-Hoo, Gramacy and Michalewicz functions. For smooth functions (Branin), the advantage of nonstationary methods (SBO and WARP) is minimal, although it is significant in the Hartmann function. For the nonstationary functions (Gramacy and Michalewicz) clearly there is an advantage of using nonstationary methods with SBO taking the lead in every test.}
  \label{fig:optbench}
\end{figure*}

All plots are based on 20 experiments using common random numbers between methods. As commented in Section \ref{sec:bo}, the number of function evaluations in each plot includes the initial design generated using \emph{latin hypercube sampling}.

\subsection{Optimization benchmarks}

\begin{figure}
  \centering
  \includegraphics[width=0.75\linewidth]{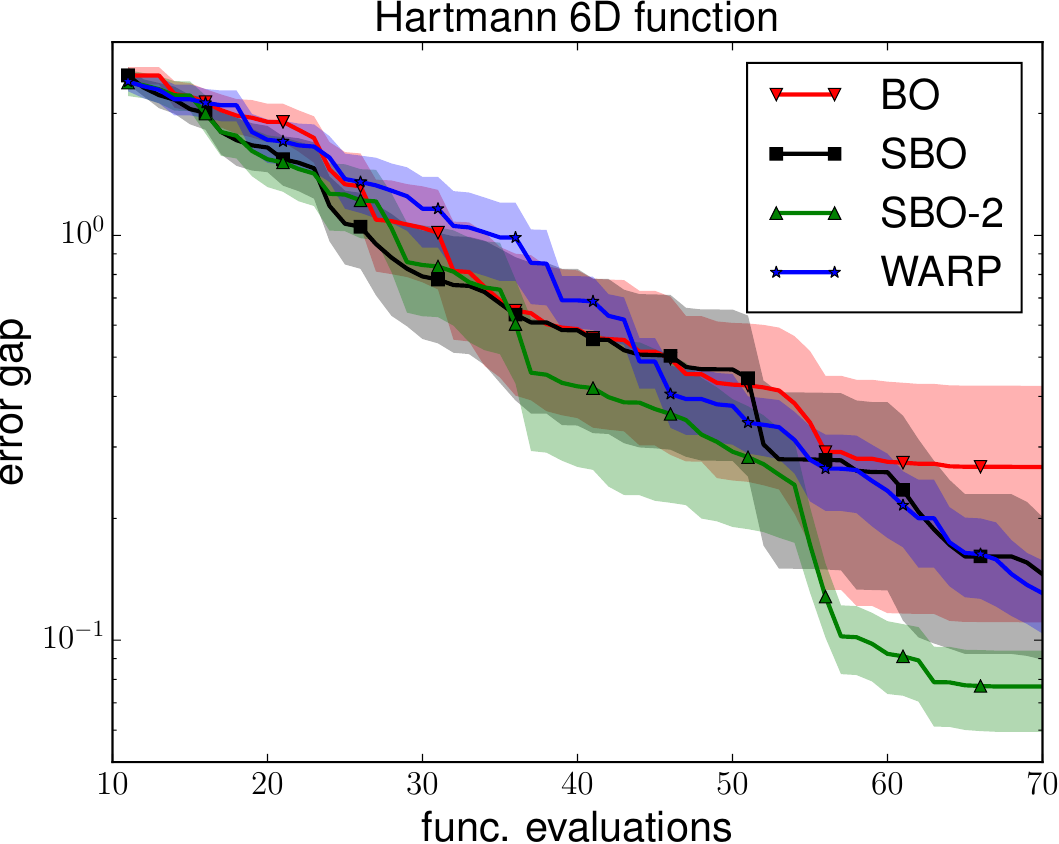}
  \caption{Hartmann 6D function results. Being a smooth function, the advantage of nonstationary methods is smaller. Still they provide better results in general. Those results are improved with a Spartan kernel with a funnel structure (local $\sigma_l = 0.05$, local $\sigma_l = 0.1$ and global) named as SBO-2.}
  \label{fig:hartlog}
\end{figure}

We evaluated the algorithms on a set of well-known test functions for global optimization both smooth or with sharp drops (see Appendix~\ref{ax:functions}). Figs. \ref{fig:optbench} and \ref{fig:hartlog} show the results for the optimization benchmarks. Clearly, functions with sharp drops or flat regions benefits from nonstationary methods, while in the case of smooth surfaces, the advantage is not as important.

Bayesian optimization is well know to behave extremely well for smooth functions with wide valleys like the Branin and Hartmann function. In this case, even plain BO preforms well in general. Therefore, there is barely room from improvement. However, even in this situation, SBO is equal or better than stationary BO. For the Branin function, there is no penalty for the use of SBO. However, the WARP method may get slower convergence due to the extra complexity. For the Hartmann function, nonstationary methods (SBO and WARP) achieve slightly better results and they are more robust as the variance on the plot is also reduced. We also use this case to show the potential of funneled kernels by adding an intermediate local kernel with $\sigma_l = 0.1$.

For nonstationary functions with sharp drops or flat regions, our method (SBO) provides excellent results. As can bee seen in Fig. \ref{fig:optbench}, in the case of the Gramacy function, it  reaches the optimum in less than 25 iterations (35 samples) for all tests. Because the function is clearly nonstationary, the WARP method outperforms standard BO, but its convergence is much slower than SBO. The Michalewicz function is known to be a hard benchmarks in global optimization. The function has a parameter to define the dimensionality $d$ and the steepness $m$. It has many local minima ($d!$) and steep drops. We used $d=10$ and $m=10$, resulting in $3628800$ minima with very steep edges. For this problem, SBO clearly outperforms the rest of the methods by a large margin.

\subsection{Machine learning hyperparameter tuning}

\begin{figure*}
  \centering
   \includegraphics[width=0.325\linewidth]{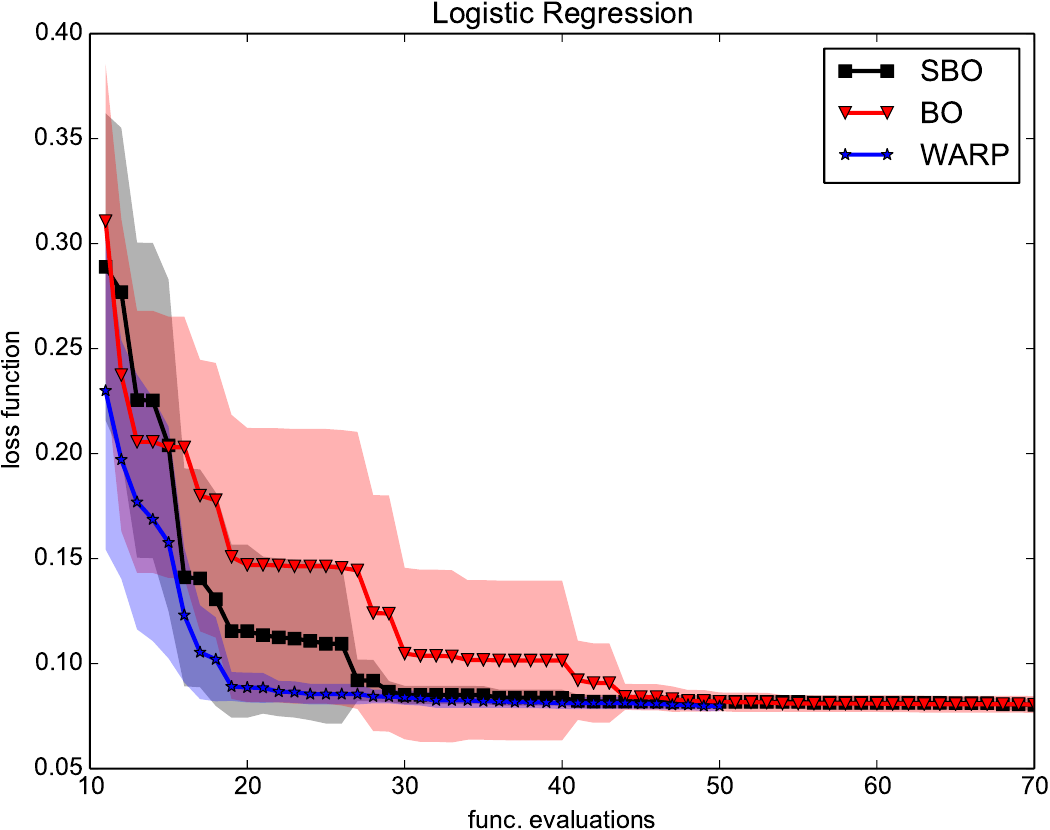} 
   \includegraphics[width=0.325\linewidth]{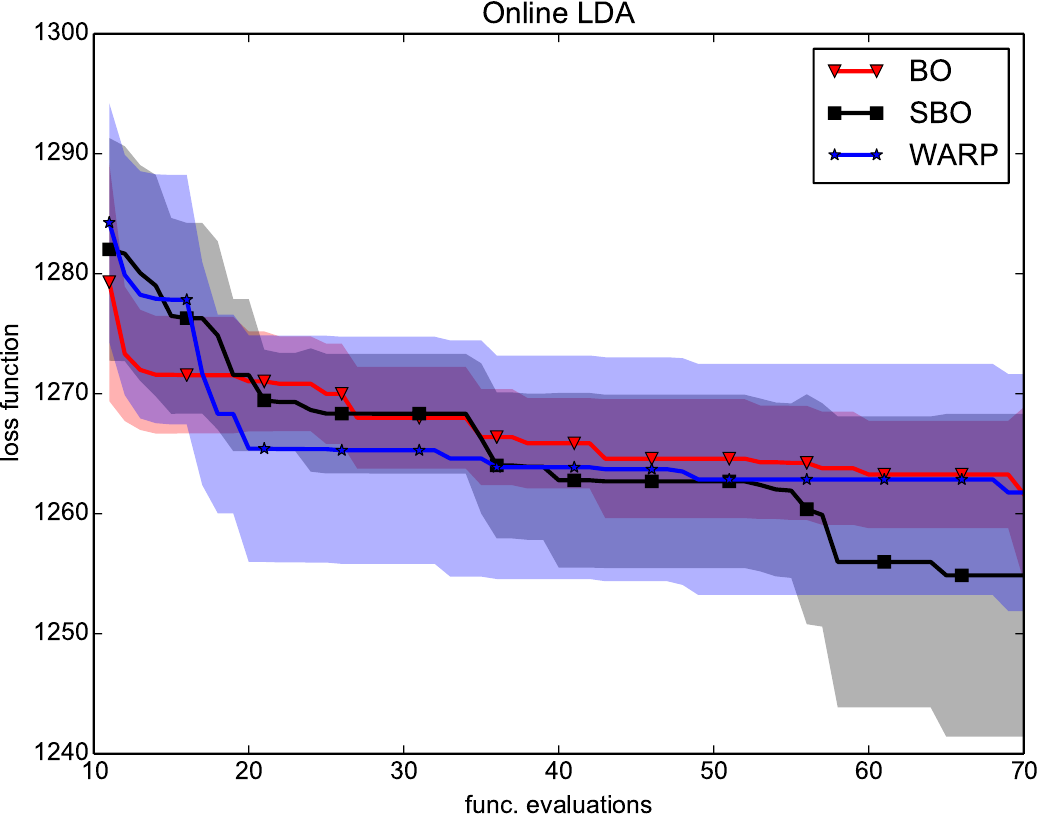}
   \includegraphics[width=0.33\linewidth]{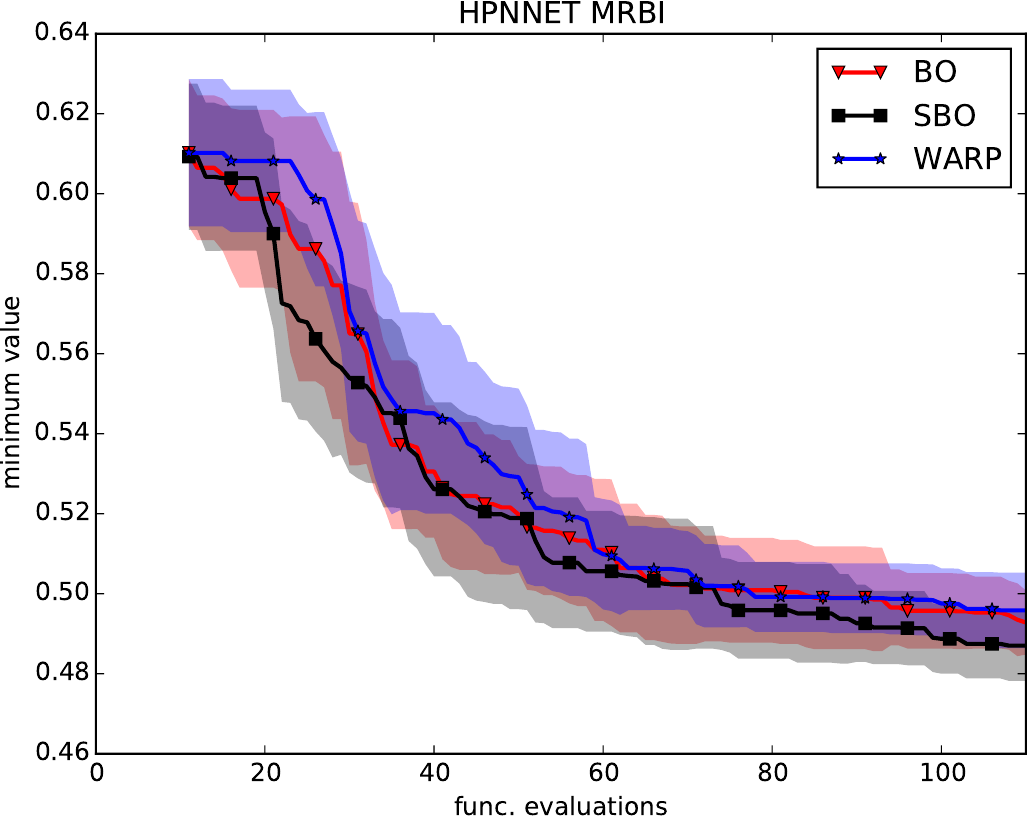} 
  \caption{Machine learning algorithm tuning. From left to right: logistic regression (4D continuous), online LDA (3D continuous), and deep neural network (HP-NNET with the MRBI dataset, 7D continuous, 7D categorical). In all cases SBO performance was the best, except for the logistic regression problem where the WARP method slightly outperformed SBO.}
  \label{fig:results}
\end{figure*}

Our next set of experiments was based on a set of problems for automatic tunning of machine learning algorithms. The results of the optimization can be seen in Fig. \ref{fig:results}.

The first problem consists on tuning the 4 parameters of a \emph{logistic regression} classifier to recognize handwritten numbers from the MNIST dataset \cite{Bergstra2011}. As can be seen in Fig. \ref{fig:results}, it is an easy problem for Bayesian optimization. Even the standard BO method were able to reach the minimum in less than 50 function evaluations. In this case, the warped method was the fastest one, with almost 20 evaluations. The proposed method had similar performance in terms of convergence, but with one order of magnitude lower execution time (see Section \ref{sec:time}). 


The second problem is based on the setup defined in Snoek et al \cite{Snoek2012} for learning topics of Wikipedia articles using \emph{online} Latent Dirichlet Allocation (LDA). It requires to tune 3 parameters. Both the standard BO and the WARP method got stuck while our method was able escape from the local minimum and outperform other methods by a large margin. Again, SBO required much lower computational cost than WARP.

Finally, we evaluated the \emph{HP-NNET} problem, based on a deep neural network written by Bergstra et al. \cite{Bergstra2011} to classify a modified MNIST dataset. In this case, the handwritten numbers were arbitrarily rotated and with random background images as distractors \cite{Bergstra2011}. The new database is called MRBI, for MNIST Rotated and with Background Images. In this case, due to the high dimensionality and heterogeneity of the input space (7 continuous + 7 categorical parameters) we tested two approaches.  First, we applied a \emph{single fully-correlated model} for all the variables. The categorical variables were mapped to integer values that were computed by rounding the query values. In this case, similarly to the Hartmann function, our method (SBO) is more precise and robust, having lower average case and smaller variance. We also tested a \emph{hierarchical model} (see Appendix~\ref{sec:hbo} for the details)

For these benchmarks a single evaluation can take hours or days of wall-time as a result of the complexity of the training process and the size of the datasets. In order to simplify the comparison and run more tests, we used the surrogate benchmarks provided by Eggensperger et al. \cite{Eggensperger2015}. They built surrogate functions of the actual behavior of the tuning process based on actual trials of the algorithms with real datasets. Each surrogate function can be evaluated in seconds, compared to the actual training time of each machine learning algorithm. The authors provide different surrogates \cite{Eggensperger2015}. We selected the Gradient Boosting as it provides the lowest prediction error (RMSE) with respect to the actual data from each problem. We explicitly rejected the Gaussian Process surrogate to avoid the advantage of perfect modeling.


\subsection{Reinforcement learning and control}
\label{sec:reinf-learn}

We evaluated SBO with several reinforcement learning problems. Reinforcement learning deals with the problem of how artificial agents perform optimal behaviors. An agent is defined by a set of variables $\s_t$ that capture the current state and configuration of the agent in the world. The agent can then perform an action $\at$ that modifies the agent state $\s_{t+1} = T(\s_t,\at)$. At each time step, the agent collects the reward signal associated with its current state and action $r_t(\s_t,\at)$. The actions are selected according to a policy function that represents the agent behavior $\atp = \pi(\s_t)$. The states, actions and transitions are modeled using probabilities to deal with uncertainty. Thus, the objective of reinforcement learning is to find the optimal policy $\pi^*$ that maximizes the expected accumulated reward
\begin{equation}
  \label{eq:optpol}
  \pi^* = \arg \max_\pi \ex_{\s_{0:T},\mathbf{a}_{1:T}}\left[\sum_{t=1}^T r_t\right]
\end{equation}
The problems we have selected are all \emph{episodic} problems with a finite time horizon. However, the methodology can be extended to infinite horizon by adding a discount factor $\gamma^t$. The expectation is evaluated using Monte Carlo, by sampling several episodes for a given policy. 

Reinforcement learning algorithms usually rely on variants of the Bellman equation to optimize the policy step by step considering each instantaneous reward $r_t$ separately. Some algorithms also rely on partial or total knowledge of the transition model $\s_{t+1} = T(\s_t,\at)$ in advance. \emph{Direct policy search} methods \cite{williams1992simple} tackle the optimization problem directly, considering equation \eqref{eq:optpol} as an stochastic optimization problem. The use of Bayesian optimization for reinforcement learning was previously called \emph{active policy search} \cite{MartinezCantin07RSS} for its connection with active learning and how samples are carefully selected based on current information.

The main advantage of using Bayesian optimization to compute the optimal policy is that it can be done with very little information. In fact, as soon as we are able to simulate scenarios and return the total reward $\sum_{t=1}^Tr_t$, we do not need to access the dynamics, the instantaneous reward or the current state of the system. There is no need for space or action discretization, building complex features or \emph{tile coding} \cite{sutton1998}. For many problems, a simple, low dimensional, controller is able to achieve state-of-the-art performance if properly optimized. 

A frequent issue for applying general purpose optimization algorithms for policy search is the occurrence of \emph{failure} states or scenarios which produces large discontinuities or flat regions due to penalties. This is opposed to the behavior of the reward near the optimal policy where small variations on a suboptimal policy can considerably change the performance achieved. Therefore, the resulting reward function presents a nonstationary behavior with respect to the policy.

\begin{figure*}
  \centering
\includegraphics[width=0.32\linewidth]{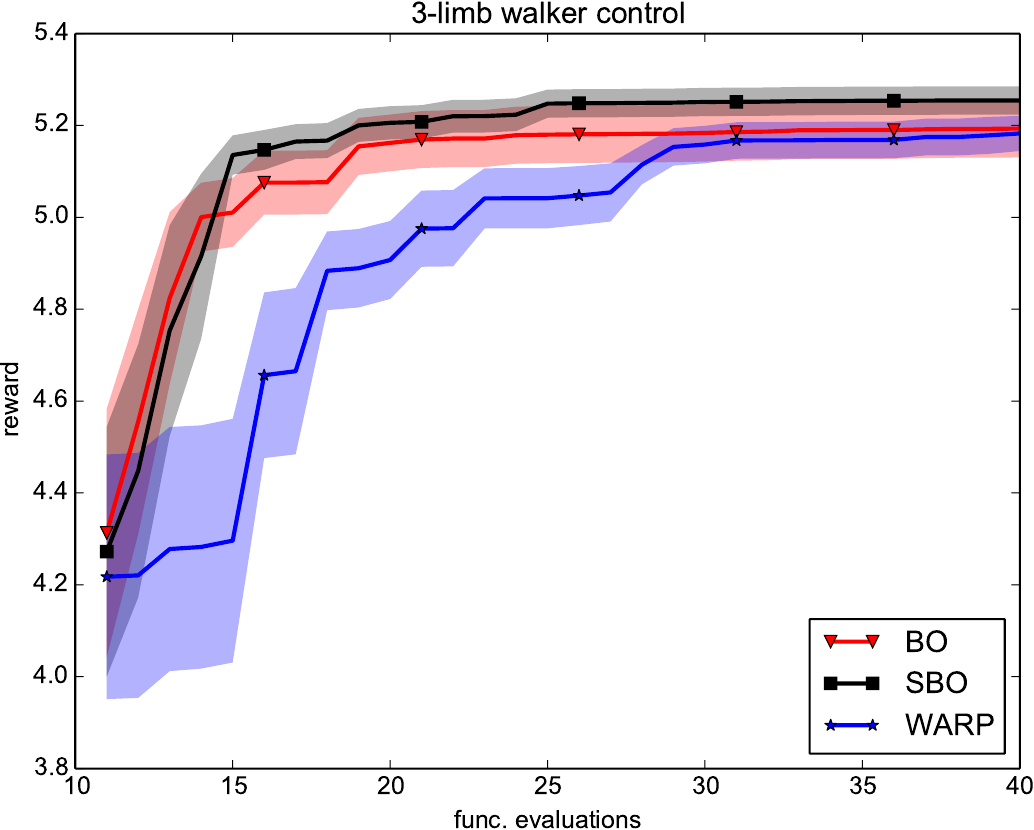}
\includegraphics[width=0.32\linewidth]{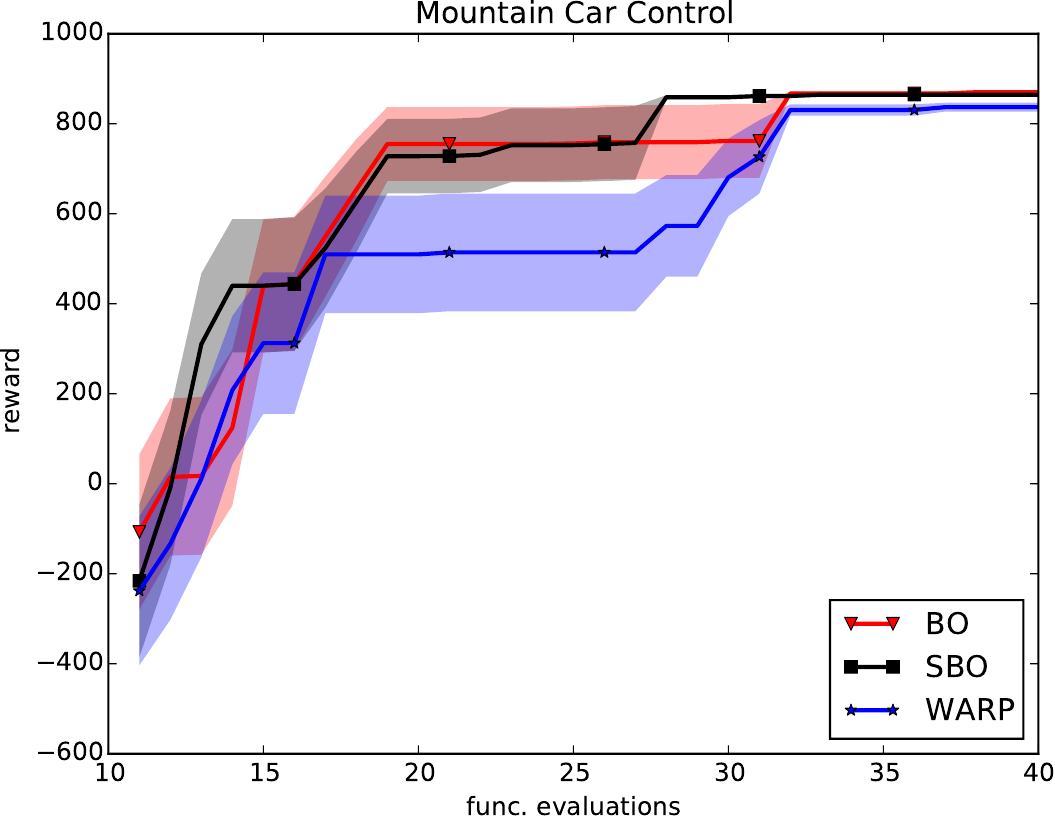}
\includegraphics[width=0.33\linewidth]{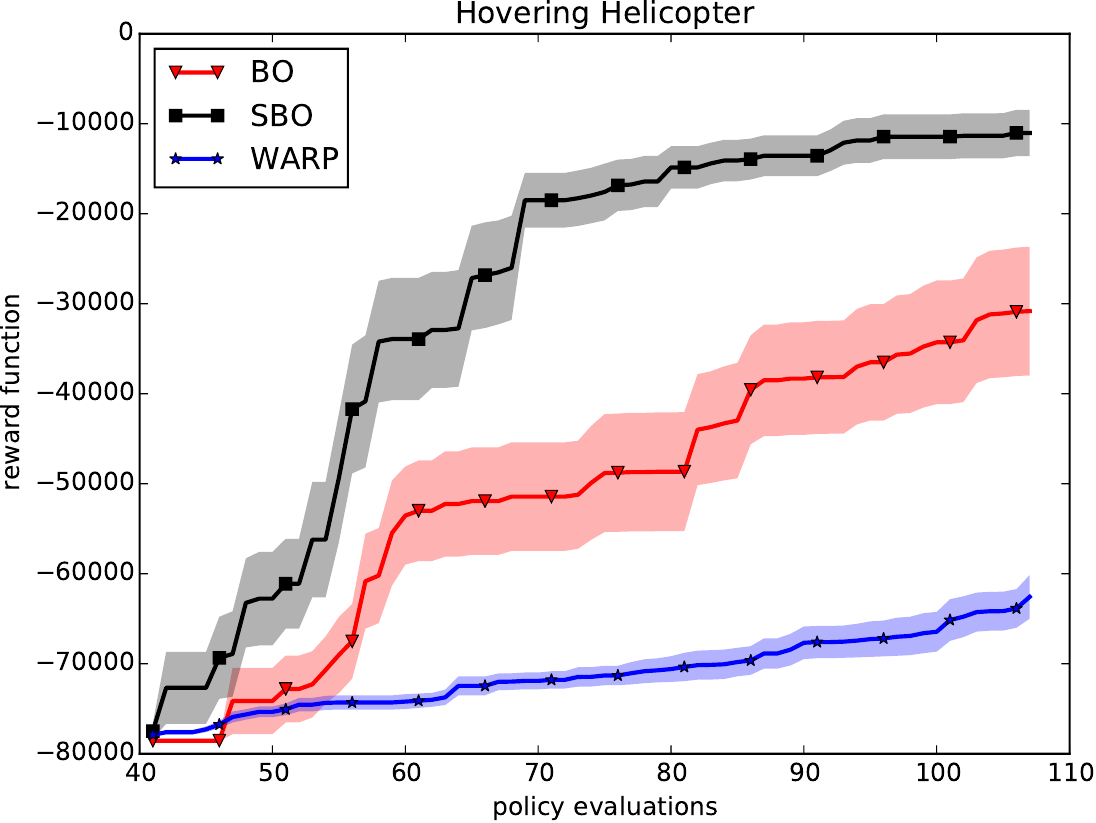}
\caption{Total reward for the three limb walker (left), the mountain car and the hovering helicopter problem. For the first problem, SBO is able to achieve higher reward, while other methods get stuck in a local maxima. For the mountain car, SBO is able to achieve maximum performance in all trials after just 27 policy trials (17 iterations + 10 initial samples). For the helicopter problem, BO and WARP have slow convergence, because many policies results in an early crash. Providing almost no information. However, SBO is able to exploit good policies and quickly improve the performance.}
  \label{fig:rlres}
\end{figure*}

We have compared our method in three well-known benchmarks with different level of complexity. The first problem is learning the controller of a three limb robot walker presented in Westervelt et al. \cite{westervelt2007feedback}. The controller modulates the walking pattern of a simple biped robot. The desired behavior is a fast upright walking pattern, the reward is based on the walking speed with a penalty for not maintaining the upright position. The dynamic controller has 8 continuous parameters. The walker problem was already used as a Bayesian optimization benchmark \cite{NIPS2014_5324}.

The second problem is the classic mountain car problem \cite{sutton1998}. We have used a simple perceptron policy with 7 parameters to compute the action based on position, speed and acceleration (see Appendix~\ref{ax:mcar}).

The third problem is the hovering helicopter from the RL-competition\footnote{http://www.rl-competition.org/}. This is one of the most challenging scenarios of the competition, being presented in all the editions. This problem is based on a simulator of the XCell Tempest aerobatic helicopter. The simulator model was learned based on actual data from the helicopter using apprenticeship learning \cite{Abbeel2006}. The model was then used to learn a policy for the real robot. The simulator included several difficult wind conditions. The state space is 12D (position, orientation, translational velocity and rotational velocity) and the action is 4D (forward-backward cyclic pitch, lateral cyclic pitch, main collective pitch and tail collective pitch). The reward is a quadratic function that penalizes both the state error (inaccuracy) and the action (energy). Each episode is run during 10 seconds (6000 control steps). If the simulator enters a terminal state (crash), a large negative reward is given, corresponding to getting the most negative reward achievable for the remaining time.

We also used a weak baseline controller that was included with the helicopter model. This weak controller is a simple linear policy with 12 parameters (weights). In theory, this controller is enough to avoid crashing but is not very robust. We show how this policy can be easily improved within few iterations. In this case, initial exploration of the parameter space is specially important because the number of policies not crashing in few control steps is very small. For most policies, the reward is the most negative reward achievable. Thus, in this case, we have used Sobol sequences for the initial samples of Bayesian optimization. These samples are deterministic, therefore we guarantee that the same number of non-crashing policies are sampled for every trial and every algorithm. We also increased the number of samples to 40.

Fig. \ref{fig:rlres} shows the performance for the three limb walker, the mountain car and the helicopter problem. In all cases, the results obtained by SBO were more efficient in terms on number of trials and accuracy, with respect to standard BO and WARP. Furthermore, we found that the results of SBO were comparable to those obtained by popular reinforcement learning solvers like SARSA \cite{sutton1998}, but with much less information and prior knowledge about the problem. For the helicopter problem, other solutions found in the literature require a larger number of scenarios/trials to achieve similar performance \cite{asbah2013reinforcement,koppejan2011neuroevolutionary}.

\subsection{Automatic wing design using a CDF software}
\label{sec:wing-design-with}

Computational fluid dynamics (CDF) software is a powerful tool for the design of mechanical and structural elements subject to interaction with fluids, such as aerodynamics, hydrodynamics or heating systems. Compared to physical design and testing in wind tunnels or artificial channels, the cost of simulation is almost negligible. Because simulated redesign is simple, CDF methods have been used for autonomous design of mechanical elements following principles from experimental design. This simulation-based autonomous design is a special case of the design and analysis of computer experiments (DACE) \cite{Sacks89SS}. Nevertheless, the computational cost of an average CDF simulation can still take days or weeks of CPU time. Therefore, the use of a sample efficient methodology like BO is mandatory. We believe this methodology has an enormous potential in the field for autonomous design. 
 
The experiment that we selected is the autonomous design of an optimal wing profile for a UAV \cite{forrester2006optimization}. We simulated a wind tunnel using the xFlow\texttrademark CDF software. The objective of the experiment was to find the shape of the wing that minimizes the drag while maintaining enough lift. We assumed a 2D simulation of the fluid along the profile of the wing. This is a reasonable and common assumption for wings with large aspect ratio (large span vs. chord), and it considerably reduces the computation time of each simulation from days to hours. For the parametrization of the profile, we used Bezier curves; however, note that Bayesian optimization is agnostic of the geometric parametrization and any other parametrization could also be used. The Bezier curve of the wing was based on 7 control points, which resulted in 14 parameters. However, adding some physical and manufacturing restrictions resulted in 5 degrees of freedom.


\begin{figure}
  \centering
  \includegraphics[width=0.45\linewidth]{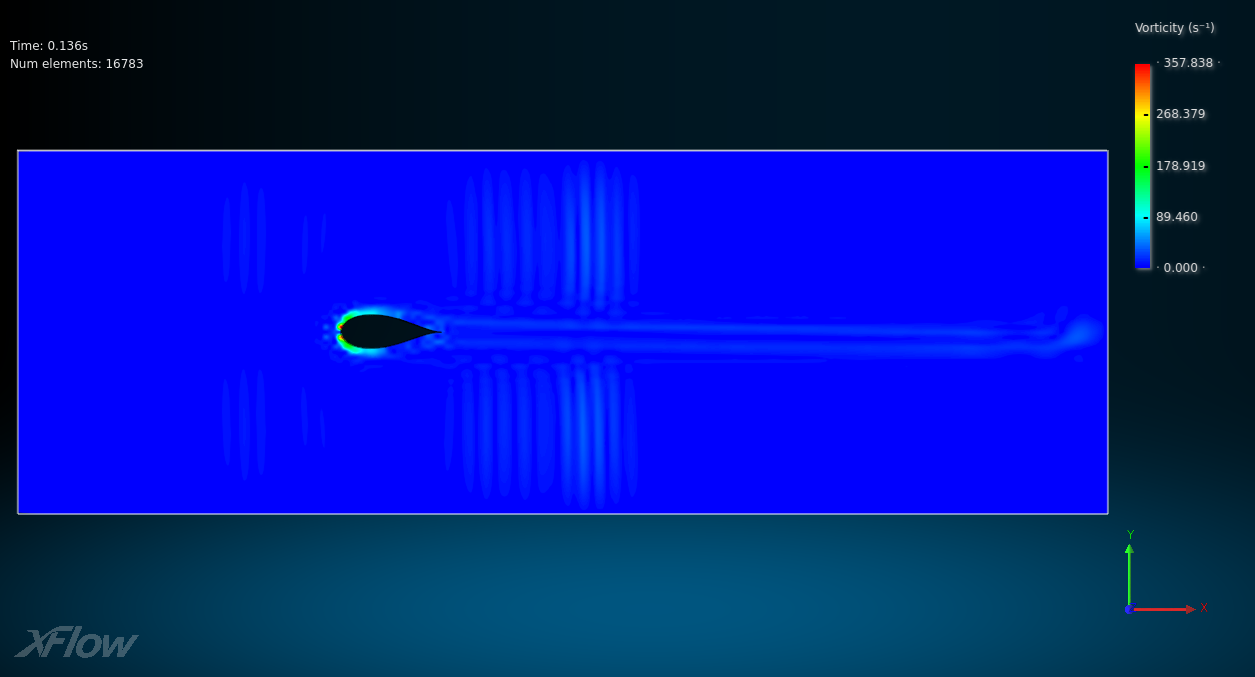}
  \includegraphics[width=0.45\linewidth]{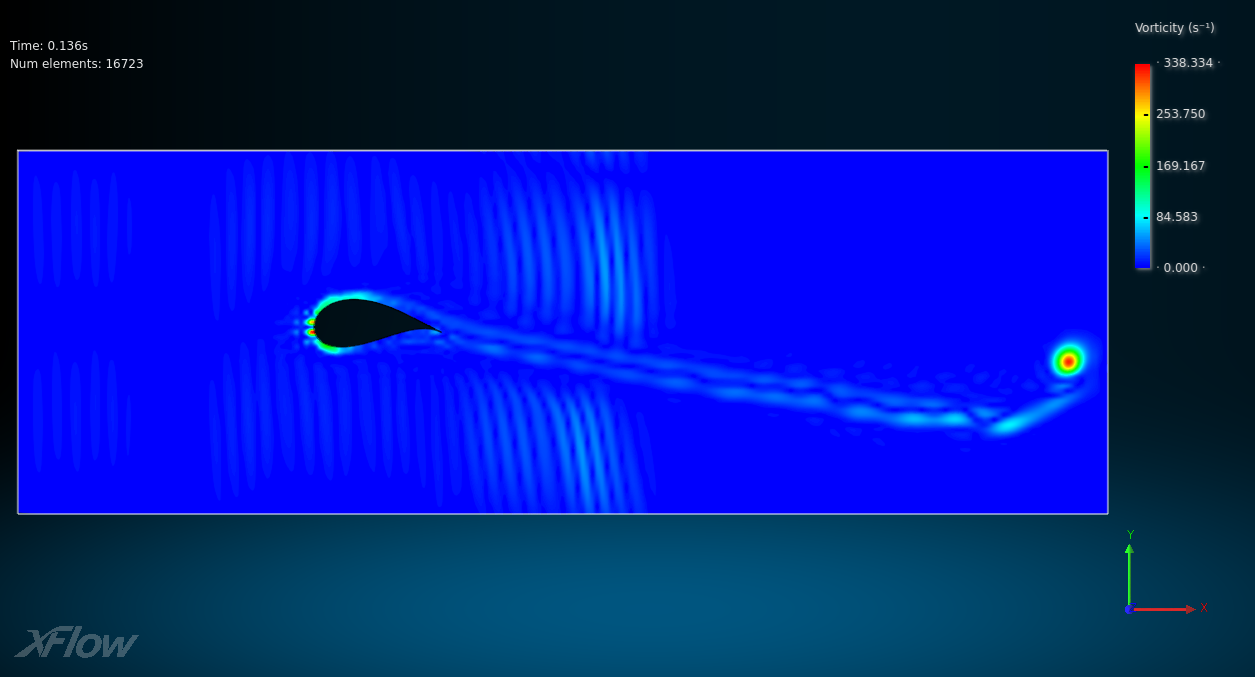}
  \caption{Vorticity plots of two different wing shapes. The left wing barely affect the trajectory of the wind, resulting in not enough lift. Meanwhile the right wing is able to provide enough lift with minimum drag.}
  \label{fig:drag}
\end{figure}

\begin{figure}
  \centering
\includegraphics[width=0.75\linewidth]{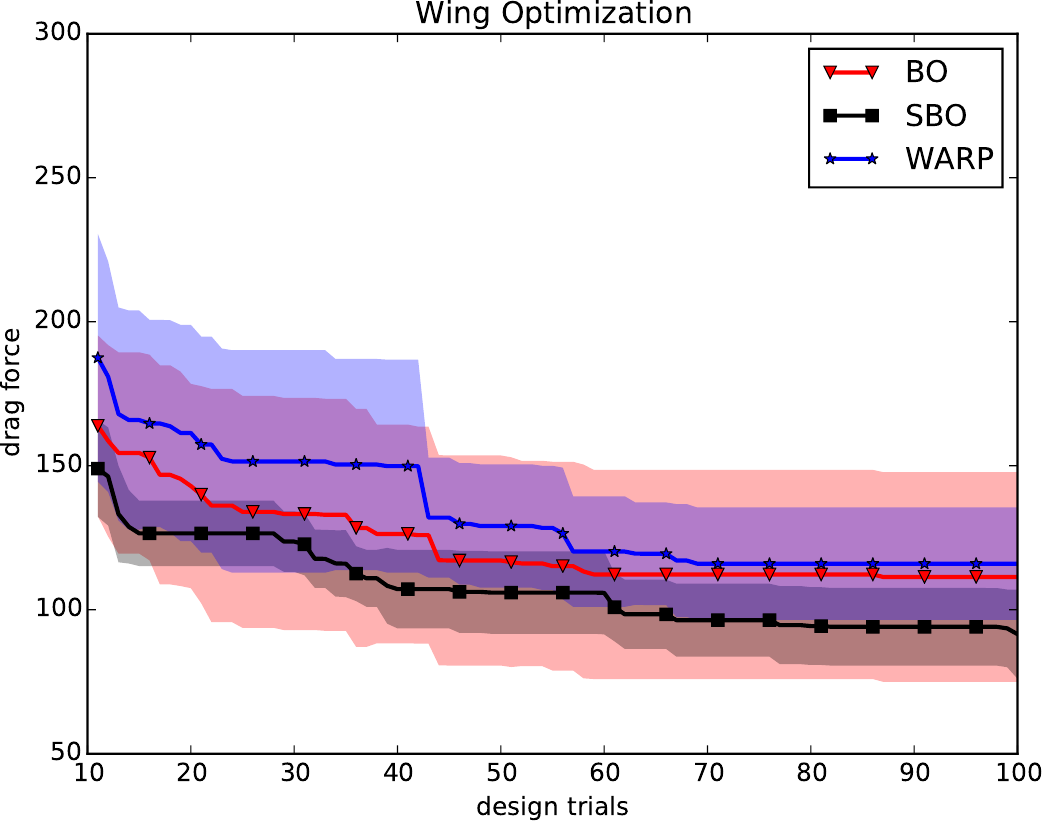}
  \caption{Results for the wing design optimization. Each plot is based on 10 runs.}
  \label{fig:wingres}
\end{figure}

\begin{table*}
  \caption{Average total CPU time in seconds.}
  \label{tab:time}
  \centering
  \begin{tabular}{|l||r|r|r|r||r|r||r|r|r|r|}
\hline
\textbf{Time (s) } & \textbf{Gramacy} & \textbf{Branin} & \textbf{Hartmann} & \textbf{Michalewicz} & \textbf{Walker} & \textbf{Mountain Car} & \textbf{ Log Reg } & \textbf{ Online LDA } & \textbf{ HPNNET }\\
\hline \hline
\#dims & 2  & 2  & 6     & 10 & 8 & 7 &  4   & 3    & 14 \\
\hline
\#evals & 60 & 40 & 70     & 210 & 40 & 40 & 50   & 70    & 110 \\
\hline\hline
  BO    & 120 & 171 & 460    & 8\,360 & 47 & 38 & 28   & 112   & 763 \\
\hline
  SBO   & 2\,481 & 3\,732 & 10\,415  & 225\,313 & 440 & 797 & 730  & 2\,131  & 28\,442 \\
\hline
  WARP  & 13\,929 & 28\,474 & 188\,942 & 4\,445\,854 & 20\,271 & 18\,972 & 9\,149 & 21\,299 & 710\,974 \\
\hline
  \end{tabular}
\end{table*}

Directly minimizing the drag presents the problem that the best solutions tends to generate flat wings that do not provide enough lift for the plane. Fluid dynamics also have chaotic properties: small changes in the conditions may produce a large variability in the outcome. For example, the flow near the trailing edge can transition from laminar to turbulent regime due to a small change in the wing shape. Thus, the resulting forces are completely different, increasing the drag and reducing the lift. Fig. \ref{fig:drag} shows comparison of a wing with no lift and the optimal design. Although there has been some recent work on Bayesian optimization with constraints \cite{gardner2014bayesian,Gelbart2014} we decided to use a simpler approach of adding a penalty with two purposes. Firts, the input space remains unconstrained, improving the performance of the optimization of the acquisition function. Second, safety is increased because points in the constrain boundary get also partly penalized as a result of GP smoothing. Under this conditions, Fig. \ref{fig:wingres} shows how both BO and WARP fail to find the optimum wing shape. However, SBO finds a better wing shape. Furthermore, it does it in few iterations.

\subsection{Computational cost}
\label{sec:time}

Table \ref{tab:time} shows the average CPU time of the different experiments for the total number of function evaluations. Due to the extensive evaluation, we had to rely on different machines for running the experiments, although all the algorithms for a single experiment were compared on the same machine. Thus, CPU time of different experiments might not be directly comparable.

The main difference between the three methods in terms of the algorithm is within the kernel function $k(\cdot,\cdot)$, which includes the evaluation of the weights in SBO and the evaluation of the warping function in WARP. The rest of the algorithm is equivalent, that is, we reused the same code.

After some profiling, we found that the time differences between the algorithms were mainly driven by the dimensionality of the hyperparameter space because MCMC was the main bottleneck. Note that, for all algorithms and experiments, we used slice sampling as recommended by the authors of WARP \cite{snoek-etal-2014a}. On one hand, the likelihood of the parameters for the Beta CDF was very narrow and the slice sampling algorithm spent many iterations before founding a valid sample. Some methods could be applied to alleviate the issues such as hybrid Monte Carlo, or sequential Monte Carlo samplers; but that remains an open problem beyond the scope of the paper. On the other hand, the evaluation of the Beta CDF was much more involved and computationally expensive than the evaluation of the Mat{\'e}rn kernel or the Gaussian weights for the Spartan kernel. That extra cost became an important factor as the kernel function was being called billions of times for each Bayesian optimization run.

It is important to note that, although Bayesian optimization is intended for expensive functions and the cost per iteration is negligible in most applications (for example: CDF simulations, deep learning algorithm tuning, etc.), the difference between methods could mean hours of CPU time for a single iteration, changing the range of potential applications. 


\section{Conclusions}

In this paper, we have presented a new algorithm called Spartan Bayesian Optimization (SBO) which combines local and global kernels in a single adaptive kernel to deal with the exploration/exploitation trade-off and the inherent nonstationarity in the search process during Bayesian optimization. We have shown that this new kernel increases the convergence speed and reduces the number of samples in many applications. For nonstationary problems, the method provides excellent results compared to standard Bayesian optimization and the state of the art method to deal with nonstationarity. Furthermore, SBO also performs well in stationary problems by improving local refinement while retaining global exploration capabilities. We evaluated the algorithm extensively in standard optimization benchmarks, automatic wing design and machine learning applications, such as hyperparameter tuning problems and classic reinforcement learning scenarios. The results have shown that SBO outperforms the state of the art in Bayesian optimization for all the experiments and tests. It requires less samples or achieves smaller optimization gap. In addition to that, we have shown how SBO was much more efficient in terms of CPU usage than other nonstationary methods for Bayesian optimization. 

The results in reinforcement learning also highlight the potential of the \emph{active policy search} paradigm for reinforcement learning. Our method is specially suitable for that paradigm. This fact opens a new opportunity for Bayesian optimization as an efficient and competitive reinforcement learning solver, without relying on the dynamics of the system, instantaneous rewards or discretization of the different spaces.

\section*{Acknowledgment}
The authors would like to thank Javier Garc\'ia-Barcos for his help on the CFD simulator and Eduardo Montijano for his valuable comments. This project has been funding in part by project DPI2015-65962-R (MINECO/FEDER, UE).




\bibliographystyle{IEEEtran}
\bibliography{IEEEabrv,optimization}

%

\begin{IEEEbiography}[{\includegraphics[width=1in,height=1.25in,clip,keepaspectratio]{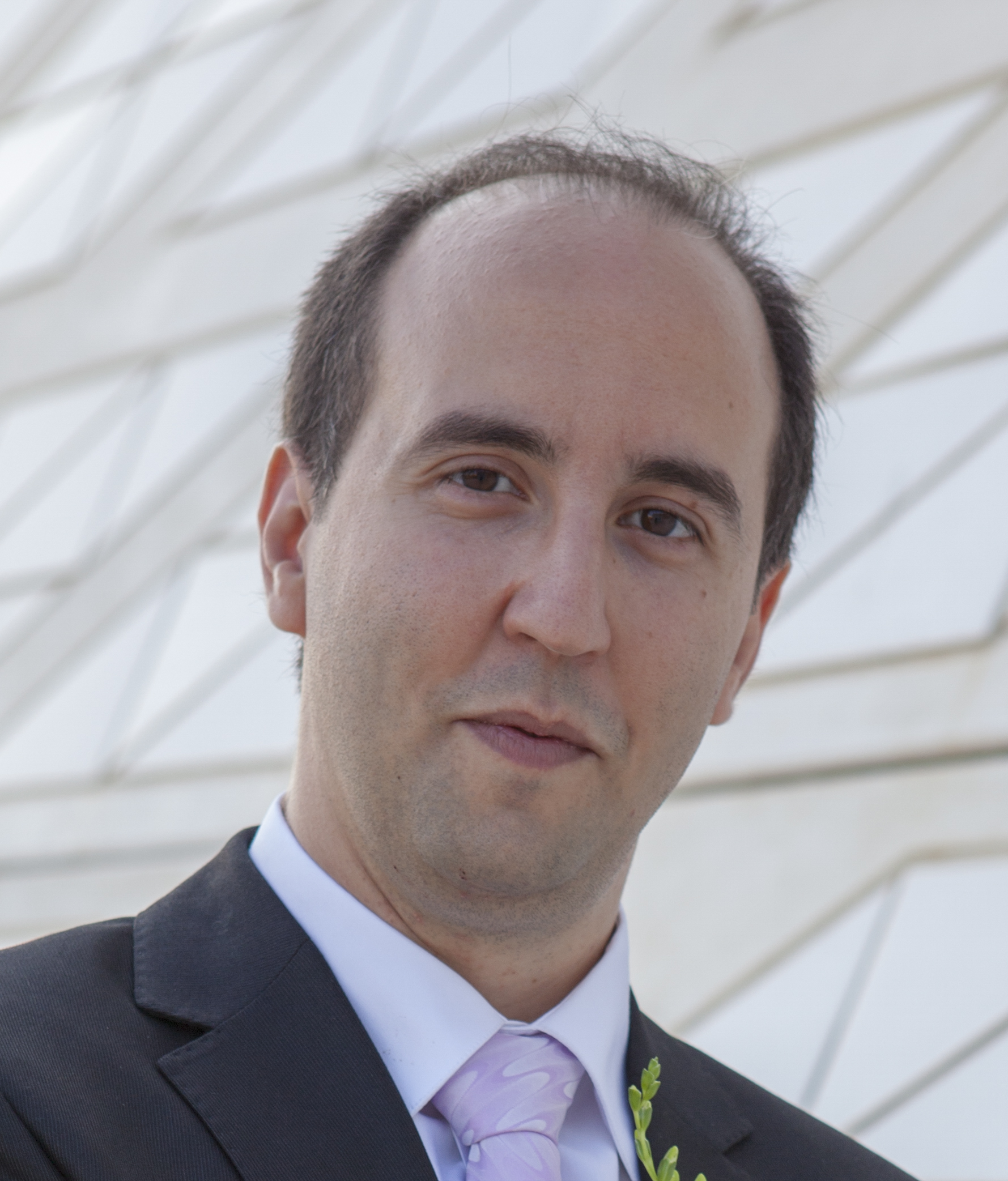}}]{Ruben Martinez-Cantin} is a senior lecturer at the Centro Universitario de la Defensa, attached to the University of Zaragora. He is also a research engineer at SigOpt, Inc. Before that, he was a postdoctoral researcher at the Insitute of Systems and Robotics at the Instituto Superior Técnico in Lisbon. Previously, he was a visiting scholar at the Laboratory of Computational Intelligence at UBC. He received a PhD and MSc in Computer Science and Electrical Engineering from the University of Zaragoza in 2008 and 2003, respectively. His research interests include Bayesian inference and optimization, machine learning, robotics and computer vision.
\end{IEEEbiography}

\appendices
\section{Origin of Spartan Bayesian optimization}
\label{ax:name}
The algorithm is called \emph{Spartan} as it follows the same intuition and analogous strategy as the Greek forces at the end of the \emph{Battle of Thermopylae}. The most likely theory claims that, the last day of the battle, a small group of forces led by spartan King Leonidas stood in the narrow pass of the Thermopylae to block the Persian cavalry, while the rest of the forces retreated to cover more terrain and avoid being surrounded by the Persian moving through a mountain path \cite{lazenby1993defence}. This dual strategy of allocate global resources sparsely while maintaining a local dense vanguard at a strategic location is emphasized within \emph{Spartan Bayesian Optimization}.

\section{Effect of kernel length-scales for Bayesian optimization}
\label{ax:length}
Like many global optimization and bandit setups, Bayesian optimization requires to control the bounds of the function space to drive exploration efficiently. In probabilistic terms, the upper and lower bounds defined by the Gaussian process in Bayesian optimization play the same role as the Lipschitz constant in classical optimization \cite{deFreitas:2013}. Wang and de Freitas \cite{wangtheoretical} pointed out that, in the case of unknown kernel hyperparameters, estimation procedures for those hyperparameters like MAP or MCMC, become overconfident as new data points become available. Thus, length-scale hyperparameters might become extremely wide, resulting in a poor exploration and slow convergence towards the optimum. This behavior is in part due to the uneven distribution of queries from Bayesian optimization. They propose adaptive bounds on the kernel hyperparameters to guarantee that the length-scale remains narrow enough to provide enough exploration. However, this approach might result in excessive global exploration when a limited budget is considered.

\section{Definitions of benchmark functions}
\label{ax:functions}
We evaluated the algorithms on a set of well-known test functions for global optimization both smooth or with sharp drops. The functions are summarized in Table \ref{tab:funcs}.

\begin{table}[h!]
  \centering
  \caption{Optimization benchmark functions}
  \begin{tabular}{|l|l|}
\hline
    \multicolumn{1}{|c|}{\textbf{Name}} & \multicolumn{1}{c|}{\textbf{Function and Domain}}\\
\hline \hline
    Branin-Hoo   & \specialcell{$f(\x) = \left(x_2 - \frac{5.1}{4\pi^2}x^2_1 + \frac{5}{\pi}x_1 - 6\right)^2$ \\[5pt] $\phantom{f(\x)} + 10\left(1-\frac{1}{8\pi}\right)\cos(x_1) + 10$ \\[10pt] $x_1 \in [-5,10]$, \quad $x_2 \in [0,15]$}\\
\hline
    Hartmann & \specialcell{$f(\x) = - \sum_{i=1}^4 \alpha_i \exp \left( -\sum_{j=1}^6 A_{ij} \left(x_j - P_{ij}\right)^2\right)$ \\[10pt] $\x \in [0, 1]^6$ \qquad see Section \ref{sec:param-hartm-6d} for $\alpha$, $A$ and $P$}\\
\hline
    Gramacy \cite{gramacy2005bayesian} & $f(\x) = x_1 \exp\left(-x_1^2-x_2^2\right)$ \qquad ${x_1, x_2} \in [-2,18]^2$ \\
\hline
    Michalewicz & \specialcell{$f(\x) = -\sum_{i=1}^{d}\sin(x_i)\sin^{2m}\left(\frac{i x_i^2}{\pi}\right)$ \\[10pt] $d=10$, $m=10$, \qquad $\x \in [0, \pi]^d$}\\
\hline
  \end{tabular}
  \label{tab:funcs}
\end{table}

\subsection{Parameters of the Hartmann 6D function}
\label{sec:param-hartm-6d}

The parameters of the Hartmann 6D function are:
\begin{align*}
\alpha &= \left(1.0, 1.2, 3.0, 3.2\right)^T ,\\
A &= \left(
    \begin{array}{cccccc}
    10 & 3 & 17 & 3.5 & 1.7 & 8\\
     0.05 & 10 & 17 & 0.1 & 8 & 14\\
     3 & 3.5 & 1.7 & 10 & 17 & 8\\
     17 & 8 & 0.05 & 10 & 0.1 & 14
    \end{array}\right), \\   
P &= 10^{-4} \left(
    \begin{array}{cccccc}
      1312 & 1696 & 5569 & 124 & 8283 & 5886\\
      2329 & 4135 & 8307 & 3736 & 1004 & 9991\\
      2348 & 1451 & 3522 & 2883 & 3047 & 6650\\
      4047 & 8828 & 8732 & 5743 & 1091 & 381
    \end{array}
\right)
\end{align*}

\section{Mountain Car}
\label{ax:mcar}

\begin{figure}
  \centering
\resizebox{.35\linewidth}{!}{%
  \begin{tikzpicture}[node distance= 2cm]
    \draw[red] (-1,-0.2) -- (-1,3);
    \draw[red] (-1,-0.2) -- (4,-0.2);
    \draw[red] (-1,2) cos (0,1);
    \draw[red] (1,0) cos (0,1);
    \draw[red] (1,0) cos (2,1);
    \draw[red] (2,1) sin (3,2);
    \draw[red] (3,2) cos (4,1.5);
    \draw[top color=white, bottom color=green!30,
           draw=green!50!black!100, drop shadow] (1,0.2) circle (2mm);
    \draw[thick,->] (3.5,3) node[anchor=south] {goal} -- (3,2.1);
    \draw[thick,->] (1.2,2) node[anchor=south] {car} -- (1,0.5);
    \draw[thick,->] (0,3) node[anchor=south] {wall} -- (-0.8,2.5);
  \end{tikzpicture}%
}
\hspace{1cm}
\resizebox{.35\linewidth}{!}{%
  \begin{tikzpicture}[node distance= 2cm]
    \node [outputpolicy] (out) {$a$};
    \node [nodepolicy, left of=out] (tanhout) {tanh};
    \node [nodepolicy, left of=tanhout] (sum) {$\Sigma$};
    \node [inputpolicy, left= 3cm of sum] (pos) {$x$};
    \node [inputpolicy, below of=pos] (vel) {$v_x$};
    \node [inputpolicy, above of=pos] (bias) {$1$};
    \node [inputpolicy, below of=vel] (acc) {$\dot{v}_x$};
    \node [nodepolicy, right of=acc] (tanh2) {tanh};

    \path [line] (tanhout) -- (out);
    \path [line] (sum) -- node[plain,anchor=south]{$w_7$} (tanhout);
    \path [line] (vel) -- node[plain,anchor=south]{$w_3$} (sum);
    \path [line] (pos) -- node[plain,anchor=south]{$w_2$} (sum);
    \path [line] (acc) -- node[plain,anchor=east]{$w_4$} (sum);
    \path [line] (bias) -- node[plain,anchor=south]{$w_1$} (sum);
    \path [line] (acc) -- node[plain,anchor=south]{$w_5$} (tanh2);
    \path [line] (tanh2) -- node[plain,anchor=east]{$w_6$} (sum);
  \end{tikzpicture}%
}

\caption{Left: Mountain car scenario. The car is underactuated and cannot climb to the goal directly. Instead it requires to move backwards to get inertia. The line in the left is an inelastic wall. Right: Policy use to control the mountain car. The inputs are the horizontal position $x_t$, velocity $v_t = x_t - x_{t-1}$ and acceleration $a_t = v_t - v_{t-1}$ of the car. The output is the car throttle bounded to $[-1, 1]$.}
  \label{fig:mcpolicy}
\end{figure}
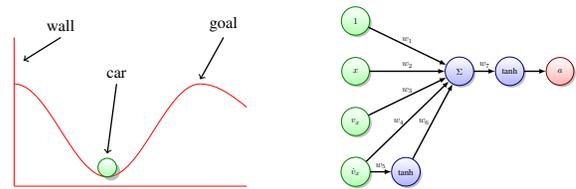

The second problem is the classic mountain car problem \cite{sutton1998}. The state of the system is the car horizontal position. The action is the horizontal acceleration $a \in [-1,1]$. Contrary to the many solutions that discretize both the state and action space, we can directly deal with continuous states and actions. The policy is a simple perceptron model inspired by Brochu et al. \cite{Brochu:2010c} as can be seen in Figure \ref{fig:mcpolicy}. The potentially unbounded policy parameters $\w = \{w_i\}_{i=1}^7$ are computed as \[\w=\tan\left(\left(\pi-\epsilon_\pi\right)\w_{01} - \frac{\pi}{2}\right)\] where $\w_{01}$ are the policy parameters bounded in the $[0,1]^7$ space. The term $\epsilon_\pi \sim 0$ was used to avoid $w_i \rightarrow \infty$.

\section{Discussion on mixed input spaces in Bayesian optimization}
\label{sec:hbo}

\begin{algorithm*}
\caption{Hierarchical Bayesian Optimization}
\label{al:hbo}
\begin{algorithmic}[1]
\State Total budget $N = N_c \cdot N_d$   \Comment{Discrete iterations times continuous iterations}
\State Split the discrete and continuous components of input space $\x = [\x^c,  \x^d]^T$
\State Initial samples for $\x^c$
\For{$n = 1 \ldots N_c$} \Comment{Outer loop}
   \State Update model (e.g.: Gaussian process) for $\x^c$
   \State Find continuous component of next query point $\x^c_{n}$ (e.g.: maximize EI)
   \State Initial samples for $\x^d$
   \For{$k = 1 \ldots N_d$} \Comment{Inner loop}
         \State Update model (e.g.: Random forest) for $\x^d \mid \x^c_{n}$
         \State Find discrete component of next query point $\x^d_{k} \mid \x^c_{n}$
         \State Combine queries and evaluate function: $y_k \gets f([\x^c_{n}, \x^d_{k}])$
   \EndFor
   \State Return optimal discrete query for current continuous query: $\x_*^{d} \mid \x^c_{n}$
\EndFor
\State Return optimal continuous query and corresponding discrete match: $\x^* = [\x_*^{c}, \x_*^{d} \mid \x_*^{c}]^T$
\end{algorithmic}
\end{algorithm*}

Although Bayesian optimization started as a method to solve classic nonlinear optimization problems with box-bounded restrictions, its sample efficiency and the flexibility of the surrogate models have attracted the interest of other communities and expanded the potential applications of the method.

In many current Bayesian optimization applications, like hyperparameter optimization, it is necessary to simultaneously optimize different kinds of input variables, for example: continuous, discrete, categorical, etc. While Gaussian processes are suitable for modeling those spaces by choosing a suitable kernel \cite{swersky2014raiders,Hutter2009}, Bayesian optimization can become quite involved as the acquisition function (criterion) must be optimized in the same mixed input space. Available implementations of Bayesian optimization like Spearmint \cite{Snoek2012} use grid sampling and rounding tricks to combine different input spaces. However, this reduces the quality of the final result compared to proper nonlinear optimization methods \cite{MartinezCantin14jmlr}. Other authors have proposed some heuristics specially designed for criterion maximization in Bayesian optimization \cite{deFreitas:2013}, but its applicability to mixed input spaces still remains an open question.

\begin{figure}
  \centering
  \includegraphics[width=0.8\linewidth]{hpnnet2-crop.pdf} 
   \includegraphics[width=0.8\linewidth]{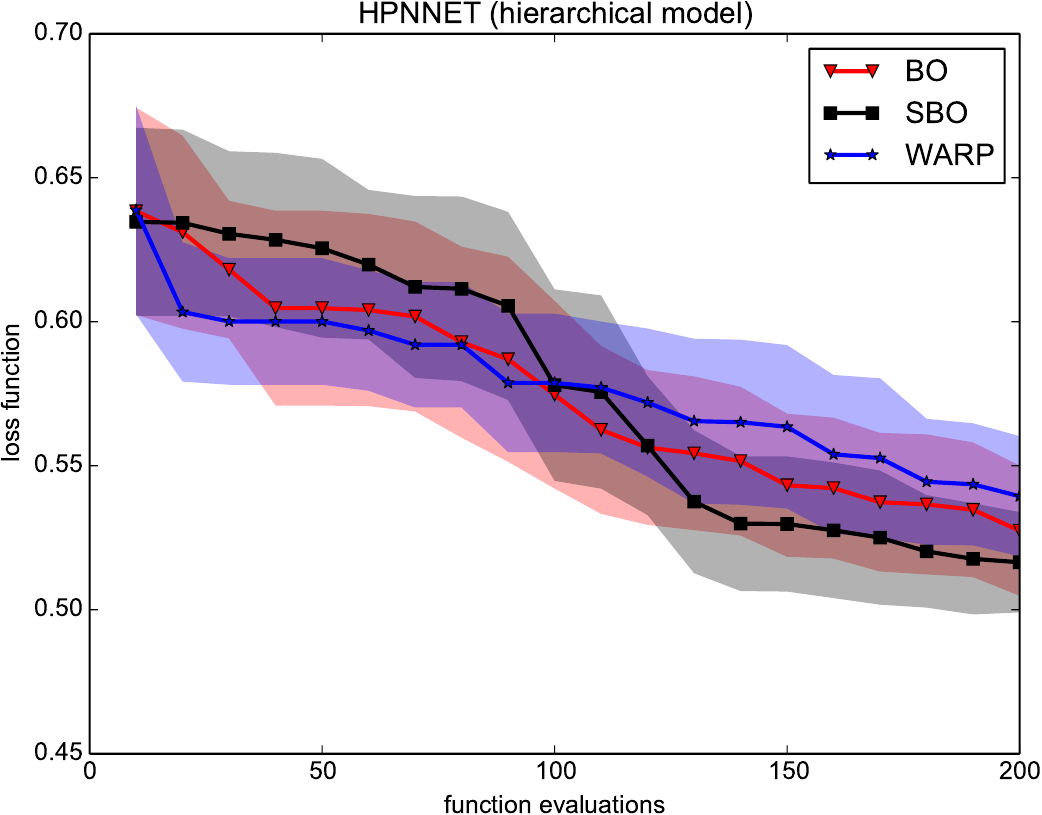} 
  \caption{Deep neural network problem using the fully correlated model (top) and the hierarchical model (bottom). Although the number of function evaluations is higher for the hierarchical model, the computational cost of the Bayesian optimization process was considerably reduced. As in the other examples, SBO performance was the best.}
  \label{fig:hierarchical}
\end{figure}

\begin{table}
  \caption{Average total CPU time in seconds.}
  \label{tab:time}
  \centering
  \begin{tabular}{|l||r|r|r|r||r|r||r|r|r|r|}
\hline
\textbf{Time (s) } & \textbf{ HPNNET } & \textbf{ HPNNET (h) }\\
\hline \hline
\#dims & 14 & 7+7 \\
\hline
\#evals & 110 & 200 (20) \\
\hline\hline
  BO    & 763 & 20 \\
\hline
  SBO   & 28\,442 & 146 \\
\hline
  WARP  & 710\,974 & 2\,853 \\
\hline
  \end{tabular}
\end{table}

We propose a \emph{hierarchical Bayesian optimization model}, where the input space $\mathcal{X}$ is partitioned between homogeneous variables, for example: continuous variables $\x^c$ and discrete variables $\x^d$. That is:
\begin{equation}
  \label{eq:sets}
\mathcal{X} = \mathcal{X}^c \cup \mathcal{X}^d \doteq \{ \x = [\x^{c},  \x^{d}]^T : \x^{c} \in \mathcal{X}^c \vee \x^{d} \in \mathcal{X}^d \}  
\end{equation}

Therefore, the evaluation of an element higher in the hierarchy implies the full optimization of the elements lower in the hierarchy. In principle, that would require many more function evaluations but, as the input space has been partitioned, the dimensionality of each separate problem is much lower. In practice, for the same number of function evaluations, the computational cost of the optimization algorithm is considerably reduced. We can also include conditional variables in the outer loop to select which computations to perform in the inner loop.

An advantage of this approach is that we can combine different surrogate models for different levels of the hierarchy. For example, using Random Forests \cite{HutHooLey11-smac} or tree-structured Parzen estimators \cite{Bergstra2011} could be more suitable as a surrogate model for certain discrete/categorical variables than Gaussian processes. We could also use specific kernels like the Hamming kernel as we used in this paper.

In contrast, we loose the correlation among variables in the inner loop, which may be counterproductive in certain situations. A similar alternative in the case where the target function is actually a combination of lower spaces, could be to use additive models, such as additive GPs \cite{kandasamy2015high}. 

For the HPNNET problem, we applied the hierarchical model to split the continuous and categorical variables in a two layer optimization process. In this case, the nonstationary algorithms (SBO or WARP) were only applied on the continuous variables (outer loop). For the categorical variables in inner loop we used a Hamming kernel  \cite{Hutter2009}:
\begin{equation}
  \label{eq:hamming}
  k_H(\x,\x' | \theta) = \exp \left(-\frac{\theta}{2}g\left(s(\x),s(\x')\right)^2\right)
\end{equation}
where $s(\cdot)$ is a function that maps continuous vectors to discrete vectors by scaling and rounding. The function $g(\cdot,\cdot)$ is defined as the Hamming distance $g(\x,\x') = |\{i:x_i \neq x'_i\}|$ so as not to impose an artificial ordering between the values of categorical parameters.

The results of the hierarchical model can be seen in Fig. \ref{fig:hierarchical}. Note that the plots are with respect to target function evaluations. However, the results of the hierarchical model are based on only 20 iterations of the outer loop, as each iteration requires 10 function evaluations in the inner loop. At early stages, SBO was trying to find a good location for the local kernel and the performance was slightly worse. However, after some data was gathered, the local kernel jumped to a good spot and the convergence was faster. Fig. \ref{fig:hierarchical} shows how the method requires more function evaluations to achieve similar results than the fully correlated approach. Table \ref{tab:time} shows the average CPU time of the different experiments for the total number of function evaluations. \emph{HPNNET (h)} is the HPNNET problem using the hierarchical model. Although it is tested with 200 function evaluations, only 20 iterations of the optimization loop are being computed. Thus, it is faster than HPNNET with a single model, which might open new applications. A similar approach has been recently proposed to deal with high dimensional problems where evaluations are not expensive \cite{ulmasov2016bayesian}.




\end{document}